\documentclass[preprint,12pt]{elsarticle}
\usepackage{amsmath,amssymb,amsthm,mathtools}
\usepackage{booktabs,array}
\usepackage{tikz}
\usepackage{hyperref}
\usepackage{enumitem}
\usepackage{microtype}
\usepackage{hyperref}
\usepackage[margin=2.0cm]{geometry}

\usepackage{tikz}
\usetikzlibrary{calc,positioning}

\tikzset{
  tsc/clauseline/.style={
    line width=0.55pt
  },
  tsc/boundary/.style={
    line width=0.9pt
  },
  tsc/literal/.style={
    font=\small,
    inner sep=1.5pt
  },
  tsc/clabel/.style={
    font=\small,
    anchor=south
  },
  tsc/annotation/.style={
    font=\small\itshape,
    fill=white,
    inner sep=1.5pt
  }
}

\journal{}

\newtheorem{theorem}{Theorem}[section]
\newtheorem{lemma}{Lemma}[section]
\newtheorem{corollary}{Corollary}[section]

\theoremstyle{definition}
\newtheorem{definition}{Definition}[section]
\newtheorem{example}{Example}[section]

\theoremstyle{remark}
\newtheorem{remark}{Remark}[section]

\newcommand{\Dlt}{\Delta}

\begin{document}
\begin{frontmatter}

\title{Automated Theorem Generation in First-Order Logic by Construction from Triangular Standard Contradictions}

\author[swjtu,lab]{Yang Xu\corref{cor1}}
\ead{xuyang@swjtu.edu.cn}
\author[swjtu,lab]{Shuwei Chen\corref{cor1}}
\ead{swchen@swjtu.edu.cn}
\author[swjtu,lab]{Hailin Guo}
\ead{guohl2021@qq.com}
\author[ulster,lab]{Jun Liu}
\ead{j.liu@ulster.ac.uk}
\cortext[cor1]{Corresponding authors. All four authors are co-first authors.}
\affiliation[swjtu]{organization={School of Mathematics, Southwest Jiaotong University},city={Chengdu},postcode={611756},country={China}}
\affiliation[ulster]{organization={School of Computing, Faculty of Computing, Engineering and the Built Environment, Ulster University},city={Belfast},state={Northern Ireland},country={United Kingdom}}
\affiliation[lab]{organization={National-Local Joint Engineering Laboratory of System Credibility Automatic Verification, Southwest Jiaotong University},city={Chengdu},postcode={611756},state={Sichuan},country={China}}

\begin{abstract}
Automated theorem generation (ATG) aims to enable machines to generate
theorems rather than merely prove given conjectures, which is a challenging problem in the field of Artificial Intelligence (AI). Existing approaches to
automated conjecturing and theorem finding commonly generate candidate
statements that must subsequently be proved or formally checked before they
can be accepted as theorems. This paper presents a constructive approach to
automated theorem generation in first-order logic based on
\emph{triangular standard contradictions}. The central idea is to construct
minimally unsatisfiable clause sets from suitably chosen first-order literals
and then generate theorems by taking the negation of one selected clause, or
of a conjunction of selected clauses, as the conclusion while retaining the
remaining clauses as premises. Minimal unsatisfiability ensures that the
remaining premise set is satisfiable, thereby preventing trivial derivation
through the principle of explosion, while the unsatisfiability of the
complete clause set guarantees the validity of the generated theorem. Thus,
the resulting theorems are correct by construction and do not require a
separate proof process for each generated theorem. Building on the previously developed $\Delta_1$ automated theorem
generator, we introduce seven additional theorem-generation methods,
$\Delta_2$--$\Delta_8$, based on triangular standard contradictions,
thereby establishing a unified family of eight methods.
Four corresponding automated theorem generators,
$\Delta_1$--$\Delta_4$, have been implemented and released.
We further develop a general $\Delta$-class theorem-generation method that
unifies these methods within a common theoretical framework. Combinatorial
analysis shows that the proposed constructions can systematically generate
large numbers of distinct theorems. The resulting framework provides a
structurally grounded approach to automated theorem generation in which
logical correctness follows intrinsically from the construction, and offers
a foundation for further work on automated reasoning, theorem discovery,
and the generation of formally guaranteed logical knowledge.
\end{abstract}

\end{frontmatter}

\section{Introduction}
\label{sec:introduction}

\noindent A theorem is a statement established from a set of axioms, definitions, or
previously established results by rigorous logical reasoning. Traditionally,
the development of a theorem follows a scientific process in which a
researcher first identifies patterns or problems from observed phenomena or
existing theories and formulates a conjecture or hypothesis---a statement
believed to be true but not yet proved. The conjecture is then abstracted
into a formal statement and subjected to rigorous proof using available
axioms, definitions, and established theorems. This process combines
creative insight with deductive rigour and often requires substantial
domain knowledge, specialised proof techniques such as proof by
contradiction and mathematical induction, and considerable human effort.

Automated conjecturing and theorem finding aim to complement this
traditional process by enabling machines to discover conjectures, lemmas,
or theorems automatically \cite{wos1993,zhang2026}. As summarised in the
recent survey by Zhang and Tan \cite{zhang2026}, existing approaches span
both symbolic and data-driven paradigms. Symbolic approaches include
top-down lemma discovery and bottom-up theory exploration based on
deductive reasoning, symbolic enumeration, generalisation, and systematic
search, whereas data-driven approaches increasingly exploit statistical
learning, neural models, and large language models (LLMs). These directions
have produced substantial progress in conjecture generation, lemma
synthesis, theory exploration, and theorem proving.

A central challenge, however, is the distinction between \emph{generating a
candidate statement} and \emph{generating a theorem}. A generated conjecture
becomes a theorem only after its validity has been established. In many
existing approaches, candidate conjectures or lemmas are therefore followed
by a separate proving, checking, or verification stage. This
generate--then--verify paradigm is natural and powerful, but its success
depends on both the quality of the generated candidates and the ability of a
subsequent prover or verifier to establish them. Recent work has also begun
to investigate theorem generation together with proof structures, further
demonstrating the growing importance of moving beyond conjecture generation
towards automated theorem finding.

This paper investigates a different route: \emph{theorem generation by
logical construction}. We develop the theory and methods of standard
contradictions introduced in our previous work \cite{xu2018,xu2026} and
study automated theorem generation (ATG) based on \emph{triangular standard
contradictions}. Standard contradictions were originally introduced for
multi-clause, dynamic and synergised automated deduction
\cite{xu2018}. Subsequent work developed constructive forms of standard
contradictions, including the standard extension, extended triangular, and
rectangular constructions \cite{xu2025ext,xu2025tri,xu2025rect}, together
with related contradiction-separation-based automated deduction methods
\cite{zeng2025,zeng2024,liu2024,liu2023a,liu2023b,cao2020,cao2021,cao2019}.

The key observation underlying the present work is that appropriate
triangular sub-standard contradictions can be constructed as
\emph{minimally unsatisfiable clause sets}. Let
\[
S=\{C_1,\ldots,C_m\}
\]
be such a clause set. Minimal unsatisfiability means that $S$ is
unsatisfiable whereas removing any clause makes the remaining clause set
satisfiable. Consequently, if one clause $C\in S$ is selected, then
\[
S\setminus\{C\}\vdash\neg C,
\]
while $S\setminus\{C\}$ itself remains satisfiable. More generally, selected
clauses can be transferred to the conclusion while the remaining clauses
form the premises. Thus, the construction provides not merely a candidate
conjecture to be passed to an independent proof search, but a theorem whose
logical validity follows automatically from the established properties of the
constructed minimally unsatisfiable clause set.

This use of minimal unsatisfiability is particularly important. If the
premises themselves were inconsistent, an arbitrary conclusion could follow
through the principle of explosion. Here, by contrast, the premises obtained
after removing the selected clause or clauses are satisfiable. The generated
theorems therefore avoid this trivial form of derivability. In this precise
sense, the proposed approach combines theorem generation with an intrinsic
logical correctness guarantee supplied by the construction itself, rather
than requiring a separate theorem-proving for every generated
theorem.

Building on the earlier $\Delta_1$ automated theorem generator
\cite{xu2025delta1}, this paper develops a systematic family of
triangular-standard-contradiction-based ATG methods. Four corresponding
generators, $\Delta_1$--$\Delta_4$, have also been released as implementations
\cite{delta1github,delta2github,delta3github,delta4github}. The principal
contributions of this paper are as follows:
\begin{itemize}
    \item We establish a unified first-order logic framework in which
    triangular standard contradictions are used to construct minimal
    unsatisfiable clause sets and, from them, logically valid theorems.

    \item We develop nine forms of theorem-generation construction,
    including the previously introduced $\Delta_1$ construction and a
    sequence of new triangular sub-standard-contradiction constructions,
    and characterise their minimal-unsatisfiability properties.

    \item We provide a general $\Delta$-class construction for generating
    minimally unsatisfiable clause sets from triangular standard
    contradictions and establish a minimum $2n$-literal construction under
    the stated conditions.

    \item We derive the number of distinct minimally unsatisfiable clause sets
    and corresponding theorems that can be generated by the proposed
    constructions, demonstrating the generative capacity of
    the framework.

    \item Unlike approaches in which a candidate conjecture is first
    generated and then independently proved or verified, the proposed
    framework generates theorems whose validity is guaranteed by the
    logical properties of the construction, while satisfiability of the
    resulting premise sets prevents trivial derivation through explosion.
\end{itemize}

The emphasis of this paper is therefore not merely on generating a large
number of theorems. Rather, it is on a constructive logical mechanism for
systematically producing distinct theorems with an intrinsic correctness
guarantee. The generated formal theorems may subsequently provide structured
logical knowledge for automated reasoning, theorem-proving benchmarks, and
potentially Neuro-Symbolic or LLM-based reasoning systems; investigation of
their domain-level interpretation and interestingness is left for future
work.

The remainder of the paper is organised as follows. Section~\ref{sec:related}
reviews closely related work and positions the proposed approach. The
following sections introduce standard contradictions and triangular standard
contradictions, respectively. We then establish the required satisfiability
results for first-order logic, present the nine theorem-generation
constructions and the general $\Delta$-class method, and analyse the number
of theorems that can be generated. The final section concludes the paper and
discusses directions for further research.

\section{Related Work}
\label{sec:related}

\subsection{Automated conjecturing, theory exploration, and theorem finding}

\noindent Automated theorem finding has long been recognised as a distinct problem in
automated reasoning. Wos \cite{wos1993} formulated the challenge of
automatically finding worthwhile theorems, rather than merely proving a
given target statement. More recently, Zhang and Tan \cite{zhang2026}
provided a comprehensive survey of automated conjecturing and theorem
finding, organising existing methods into symbolic and data-driven
approaches. The symbolic tradition includes lemma discovery, theory
exploration, deductive reasoning, enumeration, generalisation, and
systematic search, whereas recent data-driven approaches increasingly
exploit statistical learning, neural models, and large language models.

An important line of research within the symbolic tradition is
\emph{automated theory exploration}, in which a system starts from a
collection of definitions, axioms, or executable specifications and
systematically discovers conjectures and theorems describing the underlying
theory. The MATHsAiD project, for example, investigated automated
mathematical theory exploration by generating conjectures from supplied
axioms and definitions and attempting to prove them automatically
\cite{mccasland2017}. Such work illustrates the broader objective of moving
from goal-directed theorem proving towards systems capable of discovering
new mathematical knowledge.

A closely related direction concerns the automatic synthesis of useful
lemmas and conjectures. QuickSpec, introduced by Claessen et al.
\cite{claessen2010}, discovers equational conjectures about functional
programs by testing and equivalence-class reasoning. Yang et al.
\cite{yang2019} developed AdtInd for lemma synthesis in inductive reasoning,
combining generalisation and template-guided enumeration with SMT-based
counterexample filtering. Sivaraman et al. \cite{sivaraman2022} formulated
lemma synthesis as a data-driven program-synthesis problem and implemented
the resulting method as the \texttt{lfind} tactic for Coq, incorporating
candidate generation, filtering, and ranking. Puzis et al.
\cite{puzis2006} investigated the automated generation of interesting
theorems by systematically applying inference rules and filtering the
resulting consequences according to measures of interestingness.

These approaches expose two recurring challenges in automated theorem
discovery. First, a potentially very large space of candidate statements
must be generated and controlled. Second, logical validity alone does not
guarantee that a generated statement is novel, useful, difficult, or
mathematically interesting. The latter issue is conceptually distinct from
the problem studied in the present paper: our primary concern is the
systematic construction of logically valid entailments with satisfiable
premises, whereas the assessment of their domain-level usefulness or
interestingness is a separate problem.

\subsection{Learning-based and Neuro-Symbolic theorem generation}

\noindent Recent work has increasingly applied machine learning and generative models
to theorem discovery and theorem generation. Lin et al. \cite{lin2024}
introduced the ATG benchmark to evaluate whether generative language models
can generate new and reusable theorems that improve downstream theorem
proving. Their formulation emphasises an important distinction between
generating formally plausible statements and generating theorem-level
knowledge that is sufficiently useful to contribute to subsequent proofs.

REFACTOR, proposed by Zhou et al. \cite{zhou2024}, approaches theorem
discovery from existing proofs. It identifies reusable intermediate results
within proof structures and learns to extract them as standalone theorems,
which can subsequently be reused to refactor proofs and improve theorem
proving. Thus, REFACTOR discovers theorem-level knowledge already implicit
in proof data rather than constructing theorems automatically from a prescribed
logical structure.

LEMMANAID \cite{alhessi2026lemmanaid} adopts a Neuro-Symbolic approach to lemma
conjecturing. Neural generation is used to propose lemma templates, while
symbolic mechanisms instantiate them as formal conjectures. Related recent
work has investigated LLM-based mathematical reasoning and proof generation,
including inequality proving \cite{sheng2025}, while benchmarks such as
BrokenMath \cite{petrov2026brokenmath} demonstrate that plausible natural-language or
formal reasoning produced by LLMs may still exhibit systematic failures.
These developments reinforce the importance of maintaining a clear
distinction between statistical generation and formal guarantees of logical
validity.

A particularly relevant recent development is Wernhard's approach to
\emph{generating theorems by generating proof structures}
\cite{wernhard2026}. Rather than beginning with a supplied proof goal, this
method generates proof structures from a given set of axioms and obtains
associated theorems from them. It therefore goes beyond the conventional
conjecture--then--prove paradigm and is closely related to the broader
objective of direct theorem generation. The mechanism, however, is
fundamentally different from that investigated here: Wernhard's method is
centred on the enumeration and construction of proof structures, whereas
the present work constructs minimally unsatisfiable clause sets from
triangular standard contradictions and derives valid entailments from their
structural properties.

\subsection{Minimal unsatisfiability}

\noindent Minimal unsatisfiability is an important topic in propositional logic,
SAT, first-order logic, and the structural study of clause sets. A clause set $S$ is
\emph{minimally unsatisfiable} if $S$ is unsatisfiable but
\[
S\setminus\{C\}
\]
is satisfiable for every clause $C\in S$. Thus, every clause is essential
for the unsatisfiability of the complete set.

Early structural results include the work of Aharoni and Linial
\cite{aharoni1986}, who investigated minimally unsatisfiable formulas in
connection with minimal non-two-colourable hypergraphs. Subsequent research
has developed a substantial structural theory of minimally unsatisfiable
clause sets. In particular, deficiency---the difference between the number
of clauses and the number of variables---has become an important parameter
for characterising and classifying minimally unsatisfiable clause sets.
Kullmann \cite{kullmann2011} developed a systematic treatment of minimal
unsatisfiability, irredundancy, deficiency, and conflict structure for
clause sets, while later work has further investigated structural parameters
and classifications of minimally unsatisfiable clause sets
\cite{kullmann2016}.

The role of minimal unsatisfiability in the present work differs from its
usual role in SAT analysis. We do not address the problem of extracting a
minimally unsatisfiable subset from an arbitrary inconsistent formula, nor do
we primarily seek to classify minimally unsatisfiable clause sets by
deficiency. Instead, minimal unsatisfiability is used
\emph{constructively}: a triangular standard contradiction is designed so
that its associated clause set is minimally unsatisfiable. If
\[
S=\{C_1,\ldots,C_m\}
\]
is such a clause set, then for every $C_i\in S$,
\[
S\setminus\{C_i\}
\]
is satisfiable, while
\[
S\setminus\{C_i\}\models\neg C_i,
\]
that is, by the completeness theorem in first-order logic,
\[
S\setminus\{C_i\}
\vdash
\neg C_i.
\]
More generally, selected clauses can be transferred to the conclusion while
the remaining clauses form a satisfiable premise set. Minimal
unsatisfiability therefore provides the structural bridge between the standard
contradiction construction and theorem generation.

This distinction is important because unsatisfiability alone would not be
sufficient for the intended purpose. If the premise set were itself
inconsistent, an arbitrary conclusion would follow by the principle of
explosion. By starting from a minimally unsatisfiable clause set and
removing the clause or clauses transferred to the conclusion, the proposed
construction yields a satisfiable premise set. The resulting entailment is
therefore non-vacuous in the precise sense that its validity does not arise
from inconsistent premises.

\subsection{Standard contradictions and contradiction-separation deduction}

\noindent The present work also develops a separate line of research based on
\emph{standard contradictions}, a structured notion introduced in our
previous work rather than a conventional term of mathematical logic.
Xu et al. \cite{xu2018} introduced standard contradictions and
contradiction-separation deduction for multi-clause automated reasoning.
Unlike binary resolution, this framework exploits structured contradictions
involving multiple clauses and has subsequently been developed into a
family of contradiction-separation-based deduction methods
\cite{zeng2024,liu2023a,liu2023b,cao2021,cao2019}. Related studies
have investigated contradiction-separation automated theorem proving and
associated deduction mechanisms \cite{zeng2025,liu2024,cao2020}.

More recent work has focused on constructive forms of standard
contradictions. Xu et al. \cite{xu2026} systematically studied their
construction and application to multi-clause automated deduction and
theorem generation. The standard extension algorithm
\cite{xu2025ext}, the extended triangular method \cite{xu2025tri}, and the
rectangular-standard-contradiction-based theorem generator
\cite{xu2025rect} provide further constructive mechanisms for obtaining
structured contradictory clause sets.

Within this line of work, the $\Delta_1$ automated theorem generator
\cite{xu2025delta1} demonstrated that a triangular standard contradiction
can be exploited specifically for theorem generation. Implementations of
the subsequent $\Delta_1$--$\Delta_4$ generators have also been released
\cite{delta1github,delta2github,delta3github,delta4github}. The present paper
extends this development from individual constructions to a systematic
family of triangular-standard-contradiction-based constructions and a
general $\Delta$-class construction.

\subsection{Positioning of the present work}

\noindent The proposed approach is related to automated conjecturing, theory
exploration, lemma synthesis, theorem extraction, proof-structure
generation, and minimal-unsatisfiability research, but differs in the
mechanism by which theorem generation and logical validity are connected.

Representative conjecture-oriented approaches generate candidate statements
and subsequently employ testing, counterexample generation, theorem proving,
or formal checking to determine which candidates should be retained
\cite{claessen2010,yang2019,sivaraman2022}. Learning-based approaches
generate or identify potentially useful theorem-level knowledge from
statistical patterns or existing proof corpora
\cite{lin2024,zhou2024,alhessi2026lemmanaid}. Proof-structure-based theorem generation
instead constructs proof structures from which associated theorems can be
obtained \cite{wernhard2026}.

The present approach follows another route. It begins with a structurally
defined triangular standard contradiction and constructs a minimally
unsatisfiable clause set. One or more clauses are then transferred to the
conclusion, while the remaining clauses constitute the premises. The
unsatisfiability of the complete clause set guarantees the corresponding
entailment, whereas minimal unsatisfiability guarantees that the premise set
obtained by deleting an individual clause is satisfiable. For the
multi-clause forms considered later in the paper, satisfiability of the
remaining premise set is established by the corresponding construction.

The relationship among these representative paradigms can be summarised as
follows:
\[
\begin{aligned}
&\text{\textit{Conjecture-oriented approaches:}}\\[-1mm]
&\qquad
\text{generate candidate}
\rightarrow \text{prove/check}
\rightarrow \text{theorem}
\\[2mm]
&\text{\textit{Proof-structure approaches:}}\\[-1mm]
&\qquad
\text{generate proof structure}
\rightarrow \text{associated theorem}
\\[2mm]
&\text{\textit{Present approach:}}\\[-1mm]
&\qquad
\text{construct minimally unsatisfiable structure}\\[-1mm]
&\qquad\qquad
\rightarrow \text{theorem by construction}
\end{aligned}
\]

This comparison is intended to distinguish representative mechanisms rather
than to provide an exhaustive taxonomy of theorem-generation systems. The
present contribution is complementary to both symbolic theory exploration
and contemporary learning-based theorem generation. Existing approaches
address important questions such as conjecture discovery, usefulness,
interestingness, proof reuse, and learned generation. The focus here is
instead on a constructive logical mechanism that systematically produces
valid entailment from a prescribed structural condition.

This distinction also identifies a natural direction for future work.
Constructive symbolic generation can provide formal guarantees at the
logical level, whereas theory-exploration or learning-based methods may help
rank, select, interpret, or apply the resulting theorems according to
mathematical or domain-level notions of usefulness and interestingness.

\section{Standard Contradictions}
\noindent This section recalls the concept of a standard contradiction and related results~\cite{xu2026}. For standard background and content on logic and automated reasoning, see~\cite{harrison2009}.

In first-order logic, if $P$ is an $m$-ary predicate symbol and
$t_1,\ldots,t_m$ are terms, then
\[
P(t_1,\ldots,t_m)
\]
is an \emph{atomic formula}. A \emph{literal} is either an atomic formula
or its negation. Throughout this paper, $x,x_1,x_2,\ldots$ denote literals
rather than individual variables. For a literal $x$, we use $\neg x$ to
denote its \emph{complementary literal}, with double negation eliminated.
Thus, if
\[
x=P(t_1,\ldots,t_m),
\]
then
\[
\neg x=\neg P(t_1,\ldots,t_m),
\]
whereas if
\[
x=\neg P(t_1,\ldots,t_m),
\]
then
\[
\neg x=P(t_1,\ldots,t_m).
\]
Accordingly, $x$ and $\neg x$ are called \emph{complementary literals},
and the pair $\{x,\neg x\}$ is called a \emph{complementary pair}.

A \emph{clause} is a finite disjunction of literals. A finite set of
clauses
\[
S=\{C_1,\ldots,C_n\}
\]
is called a \emph{clause set} and represents the conjunction
\[
C_1\land\cdots\land C_n.
\]
As usual in first-order clausal logic, all variables occurring in a clause
are implicitly universally quantified. When set-theoretic operations on
clauses are used, a clause
\[
C=x_1\vee\cdots\vee x_m
\]
is identified with the finite set of its literals,
\[
C=\{x_1,\ldots,x_m\}.
\]

\paragraph{Clause-set and Cartesian-product notation.}
For a clause set
\[
S=\{C_1,\ldots,C_n\},
\]
we define its associated \emph{Cartesian-product structure} by
\[
\Pi(S)
:=
\prod_{i=1}^{n} C_i
=
C_1\times\cdots\times C_n.
\]
Thus, $S$ is a set of clauses, whereas $\Pi(S)$ is a set of
$n$-tuples of literals
\[
(x_1,\ldots,x_n),
\qquad x_i\in C_i, \ i=1,\ldots,n
%\right\.
\]
In particular,
\[
\Pi(S)
=
\left\{
(x_1,\ldots,x_n)
\;\middle|\;
x_i\in C_i,\ i=1,\ldots,n
\right\}.
\]

Throughout the paper, $S$ is used when referring to properties and
operations of the clause set, including satisfiability, clause deletion,
minimal unsatisfiability, and logical consequence. The notation $\Pi(S)$ is used specifically for the Cartesian-product structure underlying the
notions of standard contradiction and sub-standard contradiction.

We first recall the notion of a standard contradiction, introduced in our previous work~\cite{xu2018} as a structured representation of contradictory clause combinations, which is defined as below:

\begin{definition}[Standard contradiction~\cite{xu2018,xu2026}]
\label{def:standard-contradiction}
Let
\[
S=\{C_1,\ldots,C_n\}
\]
be a finite clause set in first-order logic, and let
\[
\Pi(S)
=
\prod_{i=1}^{n} C_i
=
C_1\times\cdots\times C_n
\]
be its associated Cartesian-product structure.
If every tuple
\[
(x_1,\ldots,x_n)\in\Pi(S),
\qquad x_i\in C_i, \ i=1,\ldots,n
%\right\,
\]
contains a complementary pair, i.e., there exist distinct indices
$i,j$ such that
\[
x_i=\neg x_j,
\]
then $\Pi(S)$ is called a \emph{standard contradiction} (SC).
Equivalently, we say that $S$ has the standard-contradiction property.
\end{definition}

\begin{corollary}[\cite{xu2018}]
\label{cor:standard-unsat}
If $\Pi(S)$ is a standard contradiction, then the associated clause set
$S$ is unsatisfiable.
\end{corollary}

\begin{example}
\label{ex:standard-contradiction}
Let
\[
S=\{C_1,C_2,C_3,C_4\},
\]
where
\[
C_1=x_1\vee x_2,\qquad
C_2=x_1\vee\neg x_2,
\]
\[
C_3=\neg x_1\vee x_2,\qquad
C_4=\neg x_1\vee\neg x_2,
\]
and $x_1,x_2$ are atomic formulas of first-order logic. Every tuple in
\[
\Pi(S)=C_1\times C_2\times C_3\times C_4
\]
contains a complementary pair. Hence $\Pi(S)$ is a standard contradiction,
and consequently $S$ is unsatisfiable by
Corollary~\ref{cor:standard-unsat}.
\end{example}

\begin{definition}[Sub-standard contradiction~\cite{xu2026,xu2025ext}]
\label{def:sub-standard}
Let
\[
S=\{C_1,\ldots,C_n\}
\]
be a clause set such that $\Pi(S)$ is a standard contradiction. For
\[
\varnothing\neq C_i^*\subseteq C_i,
\qquad i=1,\ldots,n,
\]
let
\[
S^*=\{C_1^*,\ldots,C_n^*\}.
\]
Then
\[
\Pi(S^*)
=
\prod_{i=1}^{n}C_i^*
\subseteq
\prod_{i=1}^{n}C_i
=
\Pi(S).
\]
Since every tuple in $\Pi(S^*)$ is also a tuple in $\Pi(S)$,
$\Pi(S^*)$ is itself a standard contradiction. We call
$\Pi(S^*)$ a \emph{sub-standard contradiction} of $\Pi(S)$.
\end{definition}

\begin{remark}
\label{rem:sub-standard}
The term \emph{sub-standard contradiction} refers specifically to the
componentwise Cartesian-product restriction
\[
\Pi(S^*)=\prod_{i=1}^{n}C_i^*
\subseteq
\Pi(S),
\qquad
\varnothing\neq C_i^*\subseteq C_i.
\]
It is therefore not an arbitrary subset of $\Pi(S)$. The
standard-contradiction property is hereditary under such a restriction.
\end{remark}

\begin{example}
Let
\[
S=\{C_1,C_2,C_3,C_4,C_5\},
\]
where
\begin{align*}
C_1&=x_1,\\
C_2&=x_2\vee\neg x_1,\\
C_3&=x_3\vee\neg x_2\vee\neg x_1,\\
C_4&=x_4\vee\neg x_3\vee\neg x_2\vee\neg x_1,\\
C_5&=\neg x_4\vee\neg x_3\vee\neg x_2\vee\neg x_1.
\end{align*}
It can be seen that $\Pi(S)=\prod_{i=1}^{5}C_i$ is a standard contradiction.

Let
\begin{align*}
C_1^*&=x_1,\\
C_2^*&=x_2,\\
C_3^*&=x_3\vee\neg x_2\vee\neg x_1,\\
C_4^*&=x_4\vee\neg x_3\vee\neg x_2,\\
C_5^*&=\neg x_4\vee\neg x_3\vee\neg x_2,
\end{align*}
and define
\[
S^*=\{C_1^*,C_2^*,C_3^*,C_4^*,C_5^*\}.
\]
Since
\[
\varnothing\neq C_i^*\subseteq C_i
\qquad (i=1,\ldots,5),
\]
we have
\[
\Pi(S^*)
=
\prod_{i=1}^{5}C_i^*
\subseteq
\prod_{i=1}^{5}C_i
=
\Pi(S).
\]
Hence $\Pi(S^*)$ is a sub-standard contradiction of $\Pi(S)$.
\end{example}

\section{Triangular standard contradictions}

\begin{definition}[Triangular standard contradiction~\cite{xu2026,xu2025tri}]
\label{def:triangular-standard}
Let $x_1,\ldots,x_n$ be $n$ first-order literals whose predicate symbols
are pairwise distinct and none of which is the equality predicate.
A graphical structure of the form shown in
Fig.~\ref{fig:triangular-standard-contradiction} is called a
\emph{triangular standard contradiction}. As explained below, the resulting structure is indeed
a standard contradiction.

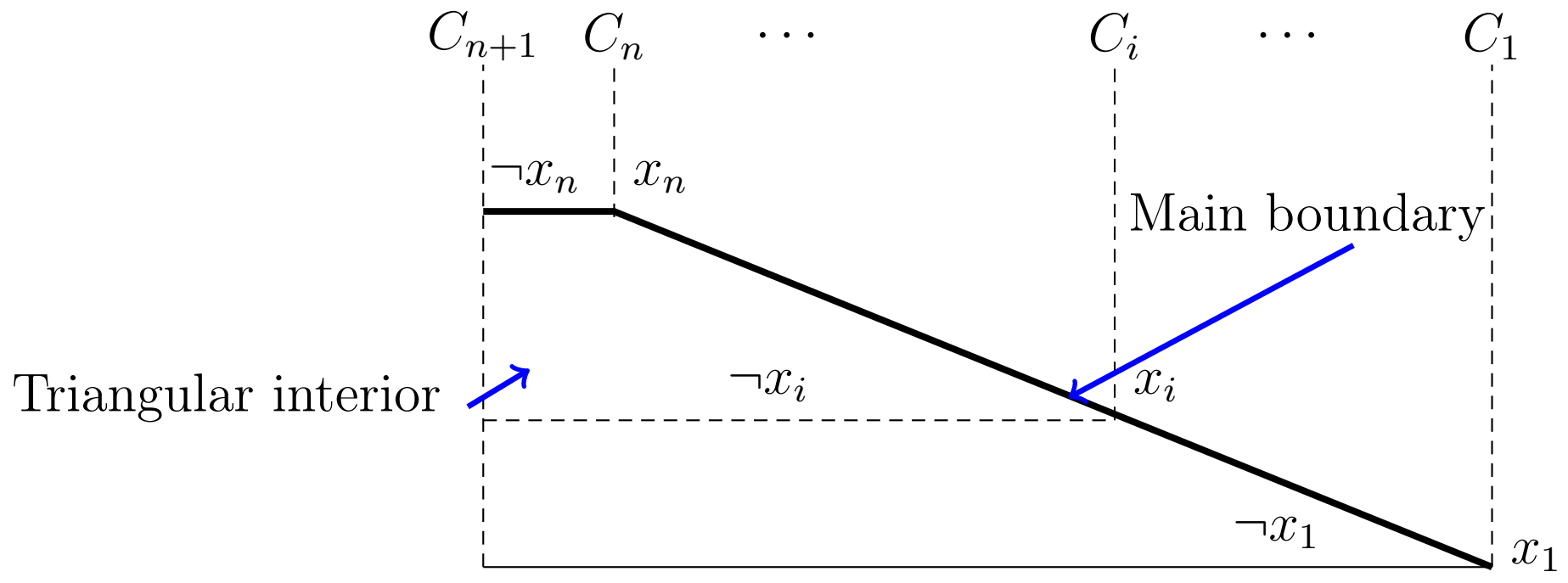
\begin{figure}[t]
\centering
\begin{tikzpicture}[x=1.20cm,y=0.88cm]

% Vertical clause lines
\draw[densely dashed] (0,0) -- (0,4.45);
\draw[densely dashed] (0.85,3.10) -- (0.85,4.45);
\draw[densely dashed] (4.10,1.30) -- (4.10,4.45);
\draw[densely dashed] (6.55,0) -- (6.55,4.45);

% Clause labels
\node at (0,4.70) {$C_{n+1}$};
\node at (0.85,4.70) {$C_n$};
\node at (2.00,4.70) {$\cdots$};
\node at (4.10,4.70) {$C_i$};
\node at (5.25,4.70) {$\cdots$};
\node at (6.55,4.70) {$C_1$};

% Main boundary
\draw[line width=1.50pt]
  (0,3.15) -- (0.85,3.15) -- (6.55,0);

% Main-boundary literals
\node[above left]  at (0.75,3.15) {$\neg x_n$};
\node[above right] at (0.85,3.15) {$x_n$};
\node[above right] at (4.10,1.30) {$x_i$};
\node[right]       at (6.55,0.10) {$x_1$};

% Representative interior literals
\node[above] at (1.85,1.30) {$\neg x_i$};
\node[above] at (5.15,0.00) {$\neg x_1$};

% Horizontal row through x_i
\draw[densely dashed] (0,1.30) -- (4.10,1.30);

% Bottom guide
\draw (0,0) -- (6.55,0);

% Main-boundary annotation
\node at (5.35,3.10) {Main boundary};
\draw[blue, ->,line width=1.25pt]
  (5.65,2.85) -- (3.80,1.50);

% Triangular-interior annotation
\node[anchor=east] at (-0.15,1.50) {Triangular interior};
\draw[blue, ->,line width=1.25pt]
  (-0.10,1.42) -- (0.30,1.75);

\end{tikzpicture}
\caption{Schematic representation of a triangular standard contradiction
$\Pi(S)$. Strictly speaking, the outer
shape shown in the figure is a trapezoid; however, since its essential
structure is triangular, we follow the established terminology and refer
to it as triangular.}
\label{fig:triangular-standard-contradiction}
\end{figure}

\vspace{0.2cm}
The graphical construction is interpreted as follows.

\begin{enumerate}[label=\arabic*)]
\item
The solid black boundary carries the first-order literals
\[
x_1,\ldots,x_n,\neg x_n
\]
and is called the \emph{main boundary} of the triangular standard
contradiction. The set
\[
\{x_1,\ldots,x_n\}
\]
is called its \emph{generating literal set}.

\item
The placement of literals in the \emph{triangular interior} obeys the following
rule. On the same horizontal row as $x_i$, if any literals other than
$x_i$ are placed in admissible interior positions, they can only be
$\neg x_i$, for
\[
i\in\{1,\ldots,n\}.
\]

\item
For each vertical line labelled $C_i$, the literals intercepted by that
line from the main boundary and the triangular interior are combined by
disjunction to form the clause $C_i$, where
\[
i=n+1,\ldots,1.
\]
The resulting clauses form the associated clause set
\[
S=\{C_{n+1},\ldots,C_1\}.
\]
Its Cartesian-product structure
\[
\Pi(S)=\prod_{i=1}^{n+1}C_i
\]
is the triangular standard contradiction represented in
Fig.~\ref{fig:triangular-standard-contradiction}.
\end{enumerate}
\end{definition}

% Insert after the definition of triangular standard contradictions.
% Requires: \usepackage{amsmath,amssymb,tikz}
% If no example environment is defined, use \newtheorem{example}{Example}[section]
% The figure is a faithful, editable TikZ reconstruction of Example.pdf.

\begin{example}[Constructing a triangular standard contradiction and its clause set]
\label{ex:triangular-table}
Let $X=\{x_1,x_2,x_3,x_4\}$ be first-order literals whose predicate
symbols are pairwise distinct and are not equality. We assume that each
$x_i$ and $\neg x_i$ are complementary literals (with the same arguments).
We illustrate how the prescribed placement of literals in a triangular
standard-contradiction diagram determines an unsatisfiable clause set.

\emph{Step 1 (Clause columns and generating boundary).}
Arrange five columns, from left to right, labelled
$C_5,C_4,C_3,C_2,C_1$. Place $\neg x_4$ in $C_5$, and
$x_4,x_3,x_2,x_1$ successively in $C_4,C_3,C_2,C_1$.
The upper red horizontal segment joins the positions of $\neg x_4$
and $x_4$, and the descending red segment passes through
$x_4,x_3,x_2,x_1$. These segments outline the upper boundary of
the triangular placement in Fig.~\ref{fig:triangular-table}.

\emph{Step 2 (Additional complementary literals).}
Place $\neg x_1$ in each of $C_5,C_3,C_2$ and $\neg x_2$ in $C_4$.
The lower red horizontal segment runs through the displayed
$\neg x_1$ positions toward $x_1$; the red vertical segments
connect the relevant positions within individual columns.
These lines are \emph{visual placement guides}: they are not
logical connectives, inference steps, or additional literal occurrences.
In particular, only the written literals, not the red line segments,
contribute to the clauses.

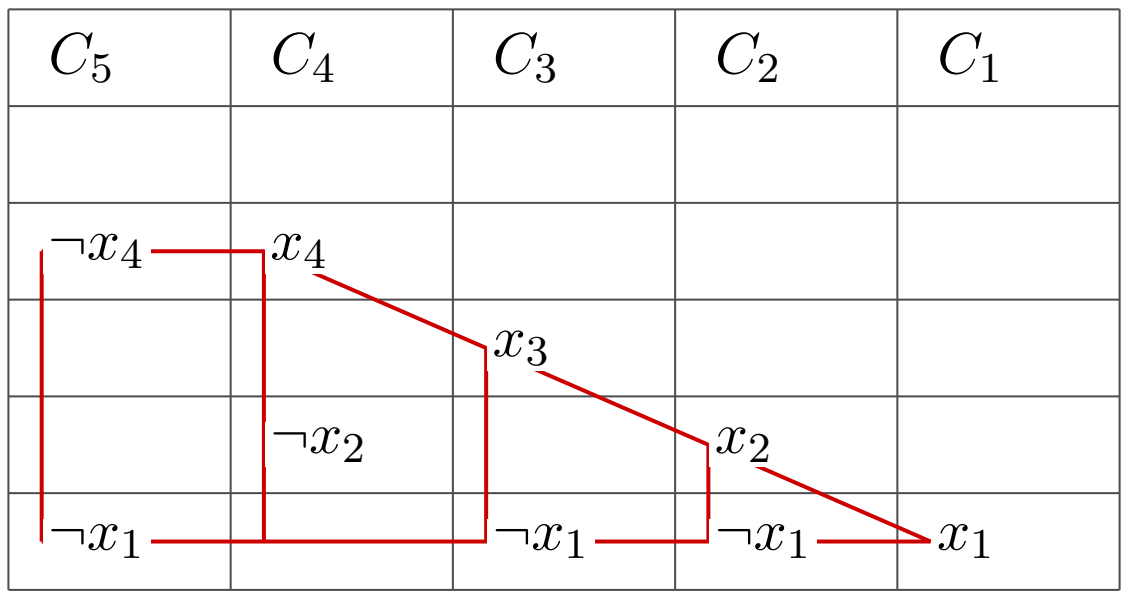
\begin{figure}[tbp]
\centering
\begin{tikzpicture}[x=1cm,y=1cm,
  every node/.style={font=\small},
  lit/.style={font=\small,inner sep=1pt,fill=white},
  gridline/.style={black!70,line width=.38pt},
  guide/.style={red!80!black,line width=.75pt}]
% Five columns, six horizontal bands (including header); grid aligned to PDF.
\def\W{1.47}
\def\H{.64}
\foreach \k in {0,...,5} {
  \draw[gridline] ({\k*\W},0) -- ({\k*\W},{6*\H});
}
\foreach \k in {0,...,6} {
  \draw[gridline] (0,{\k*\H}) -- ({5*\W},{\k*\H});
}
% Headers
\foreach \k/\j in {0/5,1/4,2/3,3/2,4/1} {
  \node[anchor=west] at ({\k*\W+.12},{5.5*\H}) {$C_{\j}$};
}
% Literal coordinates: y=3.5H x4; y=2.5H x3; y=1.5H x2; y=.5H x1.
\coordinate (nx4) at (.22,{3.5*\H});
\coordinate (x4) at ({\W+.22},{3.5*\H});
\coordinate (x3) at ({2*\W+.22},{2.5*\H});
\coordinate (x2) at ({3*\W+.22},{1.5*\H});
\coordinate (x1) at ({4*\W+.22},{.5*\H});
\coordinate (nx1a) at (.22,{.5*\H});
\coordinate (nx2) at ({\W+.22},{1.5*\H});
\coordinate (nx1c) at ({2*\W+.22},{.5*\H});
\coordinate (nx1d) at ({3*\W+.22},{.5*\H});
% Red guides, drawn first so labels remain legible.
\draw[guide] (nx4)--(x4)--(x3)--(x2)--(x1);
\draw[guide] (nx1a)--(x1);
\draw[guide] (nx4)--(nx1a);
\draw[guide] (x4)--({\W+.22},{.5*\H}); % passes through neg x_2 to bottom red boundary
\draw[guide] (x3)--(nx1c);
\draw[guide] (x2)--(nx1d);
% Literal labels
\node[lit,anchor=west] at (nx4) {$\neg x_4$};
\node[lit,anchor=west] at (x4) {$x_4$};
\node[lit,anchor=west] at (x3) {$x_3$};
\node[lit,anchor=west] at (x2) {$x_2$};
\node[lit,anchor=west] at (x1) {$x_1$};
\node[lit,anchor=west] at (nx1a) {$\neg x_1$};
\node[lit,anchor=west] at (nx2) {$\neg x_2$};
\node[lit,anchor=west] at (nx1c) {$\neg x_1$};
\node[lit,anchor=west] at (nx1d) {$\neg x_1$};
\end{tikzpicture}
\caption{A five-column triangular standard-contradiction example. The descending red line connects
the generating-literal positions; the horizontal and vertical red
segments indicate geometric placement and alignment. The lines do not
represent logical operators or additional literals. The literals in
each column form the disjuncts of the corresponding clause.}
\label{fig:triangular-table}
\end{figure}

\emph{Step 3 (Reading off the clauses).}
Each column defines a clause by disjoining precisely the literals
written in that column. Reading from left to right gives
\begin{equation}\label{eq:triangular-table-clauses}
\begin{aligned}
 C_5&=\neg x_4\vee\neg x_1, &
 C_4&=x_4\vee\neg x_2,\\
 C_3&=x_3\vee\neg x_1, &
 C_2&=x_2\vee\neg x_1,\\
 C_1&=x_1.&&
\end{aligned}
\end{equation}
Thus the associated \emph{clause set} is
\[
 S=\{C_5,C_4,C_3,C_2,C_1\},
\]
whereas its associated \emph{Cartesian-product structure} is
\[
 \Pi(S)=C_5\times C_4\times C_3\times C_2\times C_1.
\]
These are distinct mathematical objects: $S$ is a set of clauses,
while $\Pi(S)$ records their Cartesian-product structure.

\emph{Step 4 (Verifying unsatisfiability).}
We give a direct verification that does not depend on interpreting
the red lines as logical inferences. Suppose there were a model of $S$.
Since $C_1=x_1$, the literal $x_1$ would be true under every variable
assignment. The clause $C_2=x_2\vee\neg x_1$ would then force $x_2$,
and $C_4=x_4\vee\neg x_2$ would force $x_4$.
But $C_5=\neg x_4\vee\neg x_1$ would then be false, a contradiction.
Hence $S$ is unsatisfiable.

\emph{Step 5 (Unsatisfiability versus minimal unsatisfiability).}
The clause $C_3=x_3\vee\neg x_1$ is not used in the preceding
contradiction: indeed, $\{C_5,C_4,C_2,C_1\}$ is already
unsatisfiable. Therefore $S$ is \emph{not} minimally unsatisfiable.
This example illustrates the graphical construction of an
unsatisfiable clause set; the subsequent $\Delta$-type constructions
impose further structural conditions to obtain minimally
unsatisfiable clause sets, which are the basis for theorem generation
with satisfiable premises.
\end{example}

\begin{definition}[Full triangular standard contradiction%
~\cite{xu2026,xu2025tri,xu2025delta1}]
\label{def:full-triangular}
Let $\{x_1,\ldots,x_n\}$ be a generating literal set satisfying
Definition~\ref{def:triangular-standard}. Define
\[
C_1=x_1,
\]
\[
C_i
=
x_i\vee\neg x_1\vee\cdots\vee\neg x_{i-1},
\qquad i=2,\ldots,n,
\]
and
\[
C_{n+1}
=
\neg x_1\vee\neg x_2\vee\cdots\vee\neg x_n.
\]
Let
\[
S=\{C_1,\ldots,C_{n+1}\}.
\]
The Cartesian-product structure $\Pi(S)$ is called a
\emph{full triangular standard contradiction}, since all admissible
interior positions are occupied.
\end{definition}

\begin{theorem}[\cite{xu2025delta1}]
\label{thm:full-triangular-sc}
Every full triangular standard contradiction $\Pi(S)$ is a standard
contradiction.
\end{theorem}

\begin{remark}
Every triangular standard contradiction $\Pi(S^*)$ is a
sub-standard contradiction of a corresponding full triangular standard
contradiction $\Pi(S)$. Therefore,
Theorem~\ref{thm:full-triangular-sc} together with
Definition~\ref{def:sub-standard} implies that $\Pi(S^*)$ is itself a
standard contradiction. Consequently, by
Corollary~\ref{cor:standard-unsat}, its associated clause set $S^*$ is
unsatisfiable. Example ~\ref{ex:triangular-table} is, in fact, a sub-standard contradiction of a full triangular standard contradiction.
\end{remark}

\section{Satisfiability in First-Order Logic}
\begin{definition}[Minimally unsatisfiable clause set~\cite{aharoni1986}]
\label{def:minimally-unsat}
A clause set $S$ is \emph{minimally unsatisfiable} if
\begin{enumerate}[label=\arabic*)]
    \item $S$ is unsatisfiable; and
    \item for every clause $C\in S$, the clause set
    \[
    S\setminus\{C\}
    \]
    is satisfiable.
\end{enumerate}
\end{definition}

The following results are needed to establish the minimal unsatisfiability of clause sets generated by triangular standard contradictions.

\begin{theorem}[Satisfiability of a literal in first-order logic]\label{thm:literal-sat}
If an atom $x$ in first-order logic does not contain the equality predicate, then both $x$ and $\neg x$ are satisfiable.
\end{theorem}
\begin{proof}
See~\ref{app:thm41}.
\end{proof}

\begin{theorem}[Satisfiability of a Clause Set Consisting of a Conjunction of Literals and Clauses Containing These Literals as Disjuncts]\label{thm:joint-sat}
Let $x_1,x_2,\ldots,x_n$ be satisfiable literals in first-order logic. If their predicate symbols are pairwise distinct, then: (1) the set of literals $\{x_1,x_2,\ldots,x_n\}$ is jointly satisfiable; and (2) if $G_k$ is a clause in which $x_k$ occurs as one of its disjunctive literals, for $k=1,\ldots,n$, then $\{G_1,G_2,\ldots,G_n\}$ is satisfiable.
\end{theorem}
\begin{proof}
See~\ref{app:thm42}.
\end{proof}

The following result provides the general logical principle underlying the
theorem-generation methods developed in the next section.

\begin{theorem}[Theorem generation from a minimally unsatisfiable clause set]
\label{thm:mu-theorem-generation}
Let $S$ be a minimally unsatisfiable clause set, and let
\[
\varnothing\neq T\subsetneq S.
\]
Then the premise set $S\setminus T$ is satisfiable and
\[
S\setminus T
\models
\neg\bigwedge_{C\in T} C.
\]
Consequently, by the completeness theorem for first-order logic, if and only if
\[
S\setminus T
\vdash
\neg\bigwedge_{C\in T} C.
\]
Thus, the generated entailment is logically valid and has a satisfiable
premise set; in particular, the principle of explosion does not occur.
\end{theorem}
\begin{proof}
See~\ref{app:thm43}.
\end{proof}

\begin{corollary}[Single-clause theorem generation]
\label{cor:single-clause-generation}
Let $S$ be a minimally unsatisfiable clause set and let $C_{i_0}\in S$.
Then
\[
S\setminus\{C_{i_0}\}
\models
\neg C_{i_0},
\]
where $S\setminus\{C_{i_0}\}$ is satisfiable. Consequently,
\[
S\setminus\{C_{i_0}\}
\vdash
\neg C_{i_0}.
\]
\end{corollary}

\begin{remark}
\label{rem:mu-generation}
Theorem~\ref{thm:mu-theorem-generation} distinguishes two logically
separate properties of the generated result. The unsatisfiability of $S$
establishes the entailment
\[
S\setminus T
\models
\neg\bigwedge_{C\in T}C,
\]
whereas the minimal unsatisfiability of $S$ guarantees that the premise
set $S\setminus T$ is satisfiable. The latter ensures that the entailment
does not hold merely because the premises are inconsistent.
\end{remark}

\section{Theorem generation based on triangular standard contradictions}
\label{sec:theorem-generation}

\noindent As shown in~\cite{xu2025delta1}, the full triangular standard contradiction
is a special and important form of standard contradiction. Certain triangular sub-standard contradictions $\Pi(S)$ correspond to
minimally unsatisfiable clause sets $S$. By Theorem~\ref{thm:mu-theorem-generation}, for any nonempty
proper subset $T\subsetneq S$,
\[
S\setminus T
\models
\neg\bigwedge_{C\in T} C,
\]
and the premise set $S\setminus T$ is satisfiable. Hence, the entailment does not arise from the principle of explosion
(i.e., \emph{from falsehood, anything follows}), since its premise set is satisfiable. This provides the theoretical basis for automated
theorem generation from triangular standard contradictions.

\subsection{Generation of ``premises $\vdash$ conclusion'' theorems}

\noindent The minimally unsatisfiable clause sets constructed from triangular
sub-standard contradictions provide a direct basis for theorem
generation. By removing one or more clauses from such a clause set,
the remaining clauses form a satisfiable premise set, while the
removed clauses determine the conclusion. This yields the following
general procedure for generating theorems in
``premises $\vdash$ conclusion'' form.

As usual, first-order clauses are understood as universally quantified.
Accordingly, when a clause $C$ occurs as a formula, $C$ denotes its
universal closure.

\vspace{0.2cm}
The generation process is as follows.

\vspace{0.2cm}
\textbf{Step 1:}
Construct a triangular sub-standard contradiction from a generating
literal set, obtaining a corresponding minimally unsatisfiable clause
set $S$.

\textbf{Step 2:}
Suppose that $S$ contains $n+1$ clauses. Select any $k$ clauses from
$S$, say $C_1,\ldots,C_k$, where $1\leq k\leq n$.

\textbf{Step 3:}
Generate the theorem
\[
S\setminus\{C_1,\ldots,C_k\}
\ \vdash\
\neg(C_1\land\cdots\land C_k).
\]

By Theorem~\ref{thm:mu-theorem-generation}, every minimally
unsatisfiable clause set $S$ constructed by the proposed
triangular-standard-contradiction methods gives rise to
logically valid theorems with satisfiable premise sets.
Specifically, for any nonempty proper subset
$T=\{C_1,\ldots,C_k\}\subsetneq S$, we have
\[
S\setminus T \models
\neg(C_1\land\cdots\land C_k),
\]
and hence, by the completeness theorem for first-order logic,
\[
S\setminus T \vdash
\neg(C_1\land\cdots\land C_k).
\]
Thus, once the minimal unsatisfiability of a constructed
clause set has been established, theorem generation follows
directly without requiring a separate proof process
for each generated theorem.

\begin{definition}
Let $T_1:S_1\vdash C_1$ and $T_2:S_2\vdash C_2$ be two theorems. They are called the same theorem if $S_1=S_2$ and $C_1=C_2$; otherwise they are called different theorems.
\end{definition}

\begin{remark}\label{rem:multi}
Let $S$ be a minimally unsatisfiable clause set containing $n+1$
clauses. By Theorem~\ref{thm:mu-theorem-generation}, for every nonempty proper subset
\[
T=\{C_1,\ldots,C_k\}\subsetneq S,
\qquad 1\leq k\leq n,
\]
we obtain the theorem
\[
S\setminus T
\vdash
\neg(C_1\land\cdots\land C_k).
\]
Thus, considering different choices of $T$ as distinct
premise--conclusion pairs, each such clause set $S$ generates
\[
\sum_{k=1}^{n}\binom{n+1}{k}
=
2^{n+1}-2
\]
distinct theorems of this form.
\end{remark}

The central task is therefore to construct minimally unsatisfiable
clause sets from triangular sub-standard contradictions. The remainder of this section develops a family of specific
triangular-standard-contradiction constructions, denoted by
$\Delta_1,\ldots,\Delta_8$, followed by a general $\Delta$-class
construction. For each construction, the key task is to establish the
required triangular standard-contradiction structure and the minimal
unsatisfiability of its associated clause set; theorem generation then
follows from Theorem~\ref{thm:mu-theorem-generation}.

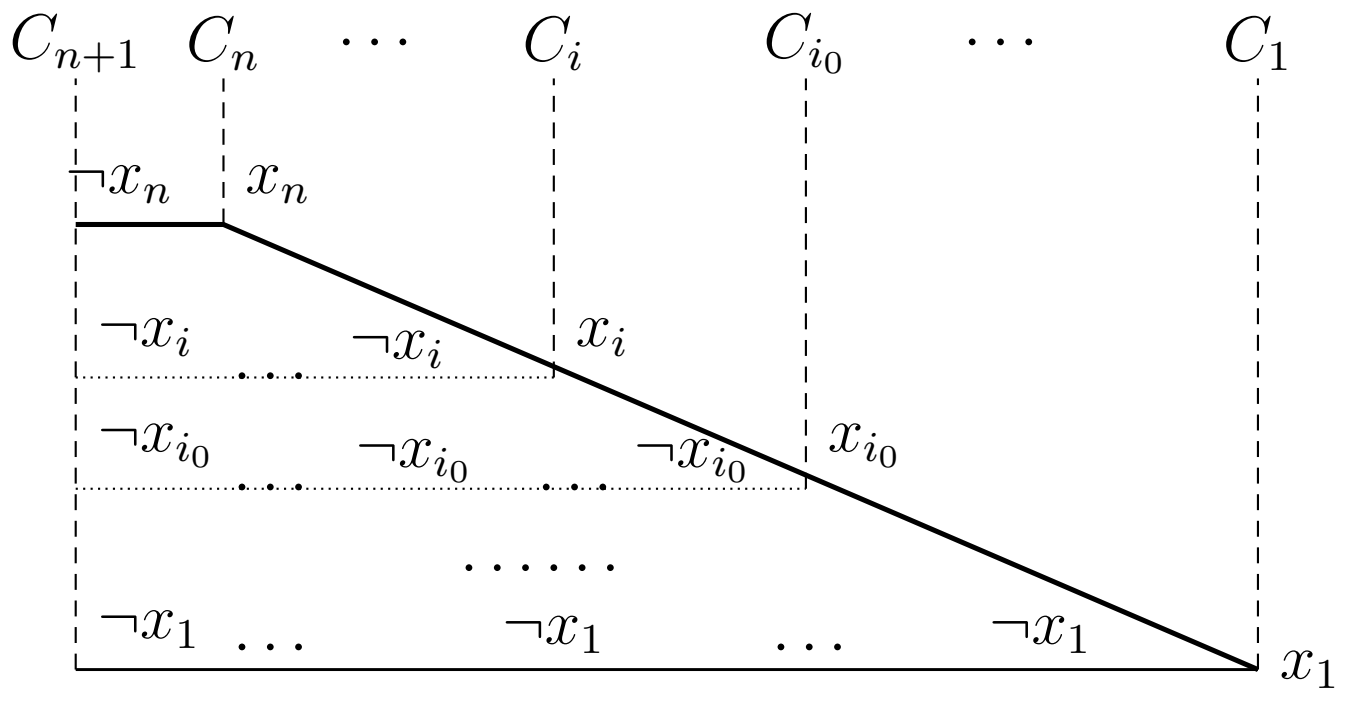
\begin{figure}[t]
\centering
\begin{tikzpicture}[x=1.15cm,y=0.92cm]

% ------------------------------------------------------------
% Key coordinates
% ------------------------------------------------------------
\coordinate (A)   at (0.00,3.20);   % neg x_n
\coordinate (Xn)  at (0.85,3.20);   % x_n
\coordinate (Xi)  at (2.75,2.10);   % x_i
\coordinate (Xi0) at (4.20,1.30);   % x_{i_0}
\coordinate (X1)  at (6.80,0.00);   % x_1

% ------------------------------------------------------------
% Clause lines
% ------------------------------------------------------------

% C_{n+1}
\draw[densely dashed]
  (0.00,0.00) -- (0.00,4.25);

% C_n
\draw[densely dashed]
  (0.85,3.20) -- (0.85,4.25);

% C_i
\draw[densely dashed]
  (2.75,2.10) -- (2.75,4.25);

% C_{i_0}
\draw[densely dashed]
  (4.20,1.30) -- (4.20,4.25);

% C_1
\draw[densely dashed]
  (6.80,0.00) -- (6.80,4.25);

% ------------------------------------------------------------
% Main boundary
% ------------------------------------------------------------
\draw[line width=0.95pt]
  (A) -- (Xn) -- (X1);

% Bottom boundary
\draw[line width=0.55pt]
  (0.00,0.00) -- (6.80,0.00);

% ------------------------------------------------------------
% Horizontal placement guides
% ------------------------------------------------------------
\draw[densely dotted]
  (0.00,2.10) -- (2.75,2.10);

\draw[densely dotted]
  (0.00,1.30) -- (4.20,1.30);

% ------------------------------------------------------------
% Clause labels
% ------------------------------------------------------------
\node at (0.00,4.50) {$C_{n+1}$};
\node at (0.85,4.50) {$C_n$};
\node at (1.75,4.50) {$\cdots$};
\node at (2.75,4.50) {$C_i$};
\node at (4.20,4.50) {$C_{i_0}$};
\node at (5.35,4.50) {$\cdots$};
\node at (6.80,4.50) {$C_1$};

% ------------------------------------------------------------
% Main-boundary literals
% ------------------------------------------------------------
\node[above]       at (0.25,3.20) {$\neg x_n$};
\node[above right] at (0.85,3.20) {$x_n$};

\node[above right] at (2.75,2.10) {$x_i$};
\node[above right] at (4.20,1.30) {$x_{i_0}$};

\node[right] at (6.80,0.00) {$x_1$};

% ------------------------------------------------------------
% Interior literals: row for neg x_i
% ------------------------------------------------------------
\node[above right] at (0.00,2.10) {$\neg x_i$};
\node at (1.15,2.10) {$\cdots$};
\node at (1.85,2.30) {$\neg x_i$};

% ------------------------------------------------------------
% Interior literals: row for neg x_{i_0}
% ------------------------------------------------------------
\node[above right] at (0.00,1.30) {$\neg x_{i_0}$};
\node at (1.15,1.30) {$\cdots$};
\node at (1.95,1.50) {$\neg x_{i_0}$};
\node at (2.90,1.30) {$\cdots$};
\node at (3.55,1.50) {$\neg x_{i_0}$};

% ------------------------------------------------------------
% Ellipsis indicating further rows
% ------------------------------------------------------------
\node at (2.70,0.72) {$\cdots\cdots$};

% ------------------------------------------------------------
% Bottom row: neg x_1
% ------------------------------------------------------------
\node[above right] at (0.00,0.00) {$\neg x_1$};
\node at (1.15,0.15) {$\cdots$};
\node at (2.75,0.25) {$\neg x_1$};
\node at (4.25,0.15) {$\cdots$};
\node at (5.55,0.25) {$\neg x_1$};

\end{tikzpicture}

\caption{Schematic representation of the full triangular
    standard contradiction underlying the $\Delta_1$ method.}
\label{fig:delta1-full-triangular}
\end{figure}

\subsection{$\Dlt_1$\protect\footnotemark: full triangular standard contradiction method}
\label{sec:delta1}

\footnotetext{The Automated Theorem Generator $\Delta_1$
is available at
\url{https://github.com/SWJTU-math/Automated_Theorem_Generator}.}

\noindent The $\Delta_1$ method~\cite{xu2025delta1} is based on a full triangular
standard contradiction $\Pi(S)$ generated by
$\{x_1,\ldots,x_n\}$, where the literals have pairwise distinct predicate
symbols and none uses equality. The corresponding construction is illustrated in
Fig.~\ref{fig:delta1-full-triangular}. Its associated clause set is
\[
S=\{C_1,\ldots,C_{n+1}\},
\]
where
\[
C_i=x_i\vee\neg x_{i-1}\vee\cdots\vee\neg x_1,
\qquad i=1,\ldots,n,
\]
with $C_1=x_1$, and
\[
C_{n+1}=\neg x_n\vee\cdots\vee\neg x_1.
\]
By~\cite{xu2025delta1} and
Theorems~\ref{thm:literal-sat}--\ref{thm:joint-sat},
$S$ is minimally unsatisfiable. 

Each ordering of the generating literals $x_1,\ldots,x_n$ on the main
boundary yields a distinct minimally unsatisfiable clause set, giving
$n!$ such clause sets. Since each clause set generates
\[
\sum_{k=1}^{n}\binom{n+1}{k}
=
2^{n+1}-2
\]
distinct theorems, the $\Dlt_1$ method generates
\[
n!\sum_{k=1}^{n}\binom{n+1}{k}
=
n!\left(2^{n+1}-2\right)
\]
mutually distinct theorems~\cite{xu2025delta1}. Each generated theorem
has a satisfiable premise set and is logically valid; hence, its
derivation does not rely on the principle of explosion. The
implementation is available as Automated Theorem Generator
$\Dlt_1$~\cite{delta1github}.

\subsection{$\Delta_2$\protect\footnotemark: Type-I triangular standard contradiction method}
\label{sec:delta2}

\footnotetext{The Automated Theorem Generator $\Delta_2$
is available at
\url{https://github.com/SWJTU-math/Automated-Theorem-Generator-2}.}

\begin{figure}[t]
\centering
\begin{tikzpicture}[x=1.05cm,y=0.92cm]

% ============================================================
% Key coordinates
% ============================================================

% Upper/main-boundary points
\coordinate (A)  at (0.00,3.30);   % neg x_n
\coordinate (B)  at (0.85,3.30);   % x_n
\coordinate (Xi) at (2.75,2.28);   % x_i
\coordinate (Xj) at (4.25,1.48);   % x_{i_0}
\coordinate (X1) at (7.00,0.00);   % x_1

% Lower-boundary points
\coordinate (Ni) at (2.00,2.05);   % neg x_i
\coordinate (Nj) at (3.55,1.30);   % neg x_{i_0}
\coordinate (N1) at (6.20,0.00);   % neg x_1

% ============================================================
% Vertical clause lines
% ============================================================
\draw[densely dashed] 
(0.00,0.00) -- (0.00,4.35);
\draw[densely dashed] 
(0.85,3.30) -- (0.85,4.35);
\draw[densely dashed] (2.75,2.28) -- (2.75,4.35);
\draw[densely dashed] 
(4.25,1.48) -- (4.25,4.35);
\draw[densely dashed]  
(7.00,0.00) -- (7.00,4.35);

% ============================================================
% Clause labels
% ============================================================
\node at (0.00,4.62) {$C_{n+1}$};
\node at (0.85,4.62) {$C_n$};
\node at (1.85,4.62) {$\cdots$};
\node at (2.75,4.62) {$C_i$};
\node at (4.25,4.62) {$C_{i_0}$};
\node at (5.45,4.62) {$\cdots$};
\node at (7.00,4.62) {$C_1$};

% ============================================================
% Hatched admissible region
% Draw this first so structural lines remain clear.
% ============================================================
\begin{scope}
\clip
  (0.00,0.00) --
  (0.00,3.30) --
  (2.00,2.05) --
  (3.55,1.30) --
  (6.20,0.00) --
  cycle;

\foreach \x in {-1.20,-0.55,...,6.00}
  \draw[blue, thin] (\x,0) -- ++(1.55,3.10);
\end{scope}

% ============================================================
% Lower boundary of admissible interior region
% neg x_n -> neg x_i -> neg x_{i_0} -> neg x_1
% ============================================================
\draw[line width=0.85pt]
  (A) -- (Ni) -- (Nj) -- (N1);

% Bottom boundary
\draw[line width=0.65pt]
  (0.00,0.00) -- (N1);

% ============================================================
% Main boundary
% neg x_n -> x_n -> x_i -> x_{i_0} -> x_1
% ============================================================
\draw[line width=0.95pt]
  (A) -- (B) -- (Xi) -- (Xj) -- (X1);

% ============================================================
% Two short connecting segments in the middle region
% ============================================================

% neg x_i -> x_i
\draw[line width=0.85pt]
  (Ni) -- (Xi);

% neg x_{i_0} -> x_{i_0}
\draw[line width=0.85pt]
  (Nj) -- (Xj);

% neg x_{1} -> x_{1}
\draw[line width=0.85pt]
  (N1) -- (X1);
% ============================================================
% Literal labels
% ============================================================

% Top pair
\node[above right] at (A) {$\neg x_n$};
\node[above right] at (B) {$x_n$};

% Middle pair i
\node[above left]  at (Ni) {$\neg x_i$};
\node[above right] at (Xi) {$x_i$};

% Middle pair i_0
\node[above left]  at (Nj) {$\neg x_{i_0}$};
\node[above right] at (Xj) {$x_{i_0}$};

% Bottom pair
\node[above left] at (N1) {$\neg x_1$};
\node[right]      at (X1) {$x_1$};

% ============================================================
% Annotation
% ============================================================
\node[align=left,anchor=east]
  at (-0.35,1.55)
  {Admissible region for\\
   interior literals according\\
   to Definition~\ref{def:triangular-standard}};

\draw[->]
  (-0.22,1.35) -- (0.25,1.55);

\end{tikzpicture}

\caption{Schematic representation of a Type-I triangular
sub-standard contradiction $\Pi(S)$ used in the $\Delta_2$ method.
The hatched region indicates the admissible region for interior literals
according to Definition~\ref{def:triangular-standard}.}
\label{fig:delta2-type-I}
\end{figure}

As illustrated in Fig.~\ref{fig:delta2-type-I}, let
\[
\{x_1,\ldots,x_n\}
\]
be a generating literal set whose literals have pairwise distinct
predicate symbols and none of which uses the equality predicate. Let
\[
S=\{C_1,\ldots,C_{n+1}\},
\]
where
\[
C_1=x_1,
\qquad
C_i=x_i\vee\neg x_{i-1}
\quad (i=2,\ldots,n),
\]
and
\[
C_{n+1}=\neg x_n.
\]
Then the Cartesian-product structure
\[
\Pi(S)=\prod_{i=1}^{n+1}C_i
\]
is a triangular sub-standard contradiction of the corresponding full
triangular standard contradiction. We call $\Pi(S)$ a
\emph{Type-I triangular standard contradiction}.

The clause set $S$ is unsatisfiable because $\Pi(S)$ is a standard
contradiction. We now show that deleting any one clause from $S$ yields
a satisfiable clause set.

Let $C_{i_0}\in S$ (${i_0}=n+1,n,\ldots,1$).

\begin{enumerate}[label=\arabic*)]
\item
If $i_0=n+1$, choose a model of
$\{x_1,\ldots,x_n\}$, which exists by
Theorems~\ref{thm:literal-sat}--\ref{thm:joint-sat}. This model satisfies every clause in
$S\setminus\{C_{n+1}\}$.
Hence, $S\setminus\{C_{n+1}\}$ is satisfiable.

\item
If $i_0=n$, choose a model of
\[
\{\neg x_n,\neg x_{n-1},\ldots,\neg x_1\},
\]
which exists by
Theorems~\ref{thm:literal-sat}--\ref{thm:joint-sat}. This model satisfies every clause in
\[
S\setminus\{C_n\}.
\]
Hence $S\setminus\{C_n\}$ is satisfiable.

\item
If
\[
i_0\in\{1,\ldots,n-1\},
\]
choose a model of
\[
\{\neg x_n,\ldots,\neg x_{i_0},
x_{i_0-1},\ldots,x_1\},
\]
which exists by
Theorems~\ref{thm:literal-sat}--\ref{thm:joint-sat}. This model satisfies every clause in
\[
S\setminus\{C_{i_0}\}
\]
is satisfied. Hence $S\setminus\{C_{i_0}\}$ is satisfiable.
\end{enumerate}

Therefore every one-clause deletion from $S$ is satisfiable. Since $S$
itself is unsatisfiable, $S$ is minimally unsatisfiable.

By placing an arbitrary nonempty set of additional literals in the
triangular interior (the region indicated by the oblique dashed line
in Fig.~\ref{fig:delta2-type-I}), in accordance with the literal-placement
principle specified in Definition~\ref{def:triangular-standard}, a Type-I triangular standard
contradiction can be extended to a new triangular standard
contradiction. The resulting structure is called a
\emph{Type-I$^*$ triangular standard contradiction}, denoted by
$\Pi(S^*)$, generated from the Type-I triangular standard
contradiction $\Pi(S)$.

\begin{remark}
\label{rem:type-I-star}
Let $S$ and $S^*$ be the clause sets associated with a Type-I
triangular standard contradiction and its corresponding
Type-I$^*$ triangular standard contradiction, respectively.
For each clause $C^*\in S^*$ and its corresponding clause $C\in S$,
we have
\[
C\subseteq C^*,
\]
where clauses are regarded as sets of literals.
Under the admissible literal-placement construction of
Definition~\ref{def:triangular-standard}, the resulting clause set $S^*$ remains
minimally unsatisfiable.
\end{remark}

\begin{figure}[t]
\centering
\begin{tikzpicture}[x=1.10cm,y=0.92cm]

% ============================================================
% Key coordinates
% ============================================================
\coordinate (A)  at (0.00,3.20);   % left/top
\coordinate (Xn) at (0.85,3.20);   % x_n
\coordinate (Xi) at (2.75,2.20);   % x_i
\coordinate (Xj) at (4.25,1.42);   % x_{i_0}
\coordinate (X1) at (7.00,0.00);   % x_1

% ============================================================
% Hatched admissible interior region
% ============================================================
\begin{scope}
  \clip
    (0.00,0.00) --
    (0.00,3.20) --
    (0.85,3.20) --
    (7.00,0.00) --
    cycle;

  \foreach \x in {-1.20,-0.55,...,6.50}
    \draw[blue, thin] (\x,0) -- ++(1.50,3.00);
\end{scope}

% ============================================================
% Main boundary
% ============================================================
\draw[line width=0.95pt]
  (0.00,3.20) -- (0.85,3.20) -- (7.00,0.00);

% middle line one
\draw[densely dashed]
  (0.00,2.20) -- (2.75,2.25);

% middle line two
\draw[densely dashed]
  (0.00,1.40) -- (4.25,1.40);
  
% Bottom boundary
\draw[line width=0.65pt]
  (0.00,0.00) -- (7.00,0.00);

% ============================================================
% Vertical clause lines
% ============================================================

% C_{n+1}
\draw[densely dashed]
(0.00,0.00) -- (0.00,4.25);

\draw[line width=1.25pt]
  (0.00,0.00) -- (0.00,3.20);
  
% C_n
\draw[densely dashed]
(0.85,3.20) -- (0.85,4.25);

% C_i
\draw[densely dashed]
  (2.75,2.20) -- (2.75,4.25);

% C_{i_0}
\draw[densely dashed]
  (4.25,1.42) -- (4.25,4.25);

% C_1
\draw[densely dashed]
  (7.00,0.00) -- (7.00,4.25);

% ============================================================
% Clause labels
% ============================================================
\node at (0.00,4.52) {$C_{n+1}$};
\node at (0.85,4.52) {$C_n$};
\node at (1.80,4.52) {$\cdots$};
\node at (2.75,4.52) {$C_i$};
\node at (4.25,4.52) {$C_{i_0}$};
\node at (5.45,4.52) {$\cdots$};
\node at (7.00,4.52) {$C_1$};

% ============================================================
% Literals on the main boundary
% ============================================================
\node[above right] at (0.85,3.20) {$x_n$};
\node[above right] at (2.75,2.20) {$x_i$};
\node[above right] at (4.25,1.42) {$x_{i_0}$};
\node[right]       at (7.00,0.00) {$x_1$};

% ============================================================
% Complementary literals on C_{n+1}
% ============================================================
\node[right] at (0.00,3.40) {$\neg x_n$};
\node[right] at (0.00,2.38) {$\neg x_i$};
\node[right] at (0.00,1.55) {$\neg x_{i_0}$};
\node[right] at (0.00,0.20) {$\neg x_1$};

% ============================================================
% Annotation
% ============================================================
\node[align=left,anchor=east]
  at (-0.38,1.55)
  {Admissible region for\\
   interior literals according\\
   to Definition~\ref{def:triangular-standard}};

\draw[->]
  (-0.32,1.32) -- (0.1,1.52);

\end{tikzpicture}

\caption{Schematic representation of a Type-II triangular
standard contradiction $\Pi(S)$ used in the $\Delta_3$ method.
The hatched region indicates the admissible region for interior literals
according to Definition~\ref{def:triangular-standard}.}
\label{fig:delta3-type-II}
\end{figure}

\subsection{$\Dlt_3$\protect\footnotemark: Type-II triangular standard-contradiction method}
\label{sec:delta3}

\footnotetext{The Automated Theorem Generator $\Dlt_3$
is available at
\url{https://github.com/SWJTU-math/Automated-Theorem-Generator-3}.}

\noindent As illustrated in Fig.~\ref{fig:delta3-type-II}, let
\[
\{x_1,\ldots,x_n\}
\]
be a generating literal set whose literals have pairwise distinct
predicate symbols and none of which uses the equality predicate. Let
\[
S=\{C_1,\ldots,C_{n+1}\},
\]
where
\[
C_i=x_i,
\qquad i=1,\ldots,n,
\]
and
\[
C_{n+1}
=
\neg x_n\vee\cdots\vee\neg x_1.
\]
The literals $\neg x_n,x_n,\ldots,x_1$ are placed along
the main boundary of the triangle. The Cartesian-product structure
\[
\Pi(S)=\prod_{i=1}^{n+1}C_i
\]
is a triangular sub-standard contradiction of the corresponding full
triangular standard contradiction. We call $\Pi(S)$ a
\emph{Type-II triangular standard contradiction}.

The clause set $S$ is unsatisfiable because $\Pi(S)$ is a standard
contradiction. We next show that deleting any clause from $S$ leaves a
satisfiable clause set.

\begin{enumerate}[label=\arabic*)]
\item
If $C_{n+1}$ is deleted, choose a model of
\[
\{x_1,\ldots,x_n\}
\]
which exists by
Theorems~\ref{thm:literal-sat}--\ref{thm:joint-sat}. This model satisfies every clause in
\[
\{C_1,\ldots,C_n\}.
\]
Hence
\[
S\setminus\{C_{n+1}\}
\]
is satisfiable.

\item
If $C_{i_0}$ is deleted for some
\[
i_0\in\{1,\ldots,n\},
\]
choose a model of
\[
\neg x_{i_0}
\]
and
\[
\{x_n,\ldots,x_{i_0+1},
x_{i_0-1},\ldots,x_1\},
\]
which exists by
Theorems~\ref{thm:literal-sat}--\ref{thm:joint-sat}. This model satisfies every clause in 
\[
\{C_{n+1}, C_n, \ldots, C_{i_0+1},
  C_{i_0-1}, \ldots, C_1\}.
\]

Hence 
\[
S\setminus\{C_{i_0}\}
\]
is satisfiable.
\end{enumerate}

Therefore every one-clause deletion from $S$ is satisfiable. Since $S$
itself is unsatisfiable, $S$ is minimally unsatisfiable.

If, starting from the Type-II construction in
Fig.~\ref{fig:delta3-type-II}, one or more admissible interior literals
are added according to the placement rule of
Definition~\ref{def:triangular-standard}, a new triangular
sub-standard contradiction $\Pi(S^*)$ is obtained. We call
$\Pi(S^*)$ a \emph{Type-II$^*$ triangular sub-standard contradiction}.

\begin{remark}
\label{rem:type-II-star}
For each corresponding clause pair $C\subseteq C^*$, every model
satisfying $C$ also satisfies $C^*$. Together with the
minimal-unsatisfiability argument established for admissible extensions,
this implies that the associated clause set $S^*$ remains minimally
unsatisfiable. 
\end{remark}

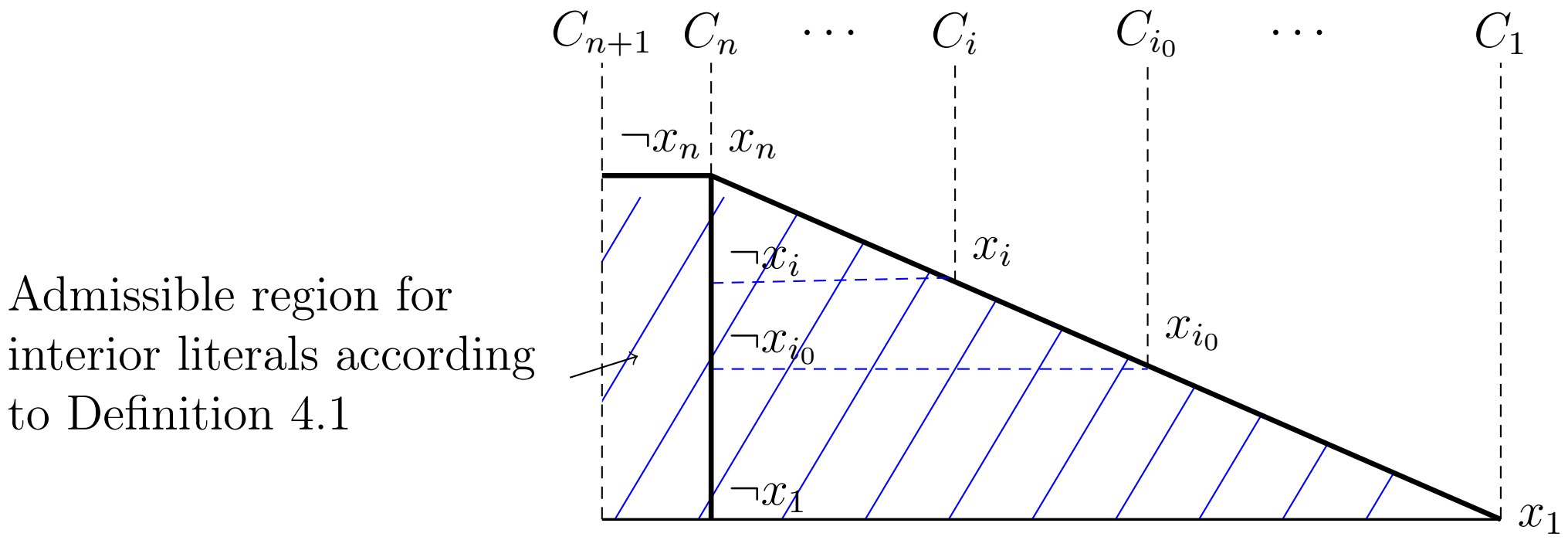
\begin{figure}[t]
\centering
\begin{tikzpicture}[x=1.10cm,y=0.92cm]

% ============================================================
% Key coordinates
% ============================================================
\coordinate (A)  at (0.00,3.20);   % neg x_n
\coordinate (Xn) at (0.85,3.20);   % x_n
\coordinate (Xi) at (2.75,2.20);   % x_i
\coordinate (Xj) at (4.25,1.42);   % x_{i_0}
\coordinate (X1) at (7.00,0.00);   % x_1

% ============================================================
% Hatched admissible interior region
% ============================================================
\begin{scope}
  \clip
    (0.00,0.00) --
    (0.00,3.20) --
    (0.85,3.20) --
    (7.00,0.00) --
    cycle;

  \foreach \x in {-1.20,-0.55,...,6.50}
    \draw[blue, thin] (\x,0) -- ++(1.50,3.00);
\end{scope}

% ============================================================
% Main boundary
% ============================================================
\draw[line width=1.25pt]
  (A) -- (Xn) -- (X1);

% middle line one
\draw[blue, densely dashed]
  (0.85,2.20) -- (2.75,2.25);

% middle line two
\draw[blue, densely dashed]
  (0.85,1.40) -- (4.25,1.40);
  
% Bottom boundary
\draw[line width=0.65pt]
  (0.00,0.00) -- (7.00,0.00);

==============================
% Vertical clause lines
% ============================================================

% C_{n+1}
\draw[densely dashed]
(0.00,0.00) -- (0.00,4.25);

% C_n -- extends through the triangular interior

% C_n
\draw[densely dashed]
(0.85,3.20) -- (0.85,4.25);

\draw[line width=1.25pt]
  (0.85,0.00) -- (0.85,3.20);
  
% C_i
\draw[densely dashed]
  (2.75,2.20) -- (2.75,4.25);

% C_{i_0}
\draw[densely dashed]
  (4.25,1.42) -- (4.25,4.25);

% C_1
\draw[densely dashed]
  (7.00,0.00) -- (7.00,4.25);

% ============================================================
% Clause labels
% ============================================================
\node at (0.00,4.52) {$C_{n+1}$};
\node at (0.85,4.52) {$C_n$};
\node at (1.80,4.52) {$\cdots$};
\node at (2.75,4.52) {$C_i$};
\node at (4.25,4.52) {$C_{i_0}$};
\node at (5.45,4.52) {$\cdots$};
\node at (7.00,4.52) {$C_1$};

% ============================================================
% Main-boundary literals
% ============================================================
\node[above right] at (A)  {$\neg x_n$};
\node[above right] at (Xn) {$x_n$};
\node[above right] at (Xi) {$x_i$};
\node[above right] at (Xj) {$x_{i_0}$};
\node[right]       at (X1) {$x_1$};

% ============================================================
% Complementary literals belonging to C_n
% ============================================================
\node[right] at (0.85,2.40) {$\neg x_i$};
\node[right] at (0.85,1.60) {$\neg x_{i_0}$};
\node[right] at (0.85,0.20) {$\neg x_1$};

% ============================================================
% Annotation
% ============================================================
\node[align=left,anchor=east]
  at (-0.38,1.55)
  {Admissible region for\\
   interior literals according\\
   to Definition~\ref{def:triangular-standard}};

\draw[->]
  (-0.25,1.32) -- (0.28,1.52);

\end{tikzpicture}

\caption{Schematic representation of a Type-III triangular
standard contradiction $\Pi(S)$ used in the $\Delta_4$ method.
The hatched region indicates the admissible region for interior literals
according to Definition~\ref{def:triangular-standard}.}
\label{fig:delta4-type-III}
\end{figure}

\subsection{$\Dlt_4$\protect\footnotemark: Type-III triangular standard-contradiction method}
\label{sec:delta4}

\footnotetext{The Automated Theorem Generator $\Dlt_4$
is available at
\url{https://github.com/SWJTU-math/Automated-Theorem-Generator-4}.}

\noindent As illustrated in Fig.~\ref{fig:delta4-type-III}, let
\[
\{x_1,\ldots,x_n\}
\]
be a generating literal set whose literals have pairwise distinct
predicate symbols and none of which uses the equality predicate. Let
\[
S=\{C_1,\ldots,C_{n+1}\},
\]
where
\[
C_{n+1}=\neg x_n,
\]
\[
C_n=x_n\vee\neg x_{n-1}\vee\cdots\vee\neg x_1,
\]
and
\[
C_i=x_i,
\qquad i=1,\ldots,n-1.
\]
The literals $\neg x_n,x_n,\ldots,x_1$ are placed along
the main boundary of the triangle. The Cartesian-product structure
\[
\Pi(S)=\prod_{i=1}^{n+1}C_i
\]
is a triangular sub-standard contradiction of the corresponding full
triangular standard contradiction. We call $\Pi(S)$ a
\emph{Type-III triangular standard contradiction}.

The clause set $S$ is unsatisfiable because $\Pi(S)$ is a standard
contradiction. We now show that deleting any one clause from $S$ yields
a satisfiable clause set.

Let $C_{i_0}\in S$ (${i_0}=n+1,n,\ldots,1$). According to Theorems~\ref{thm:literal-sat}--\ref{thm:joint-sat}, we have

\begin{enumerate}[label=\arabic*)]
\item
If $i_0=n+1$, choose a model of
\[
\{x_n,\ldots,x_1\},
\]
which exists by
Theorems~\ref{thm:literal-sat}--\ref{thm:joint-sat}. This model satisfies every clause in
\[
\{C_n,\ldots,C_1\}.
\]
Hence
\[
S\setminus\{C_{n+1}\}
\]
is satisfiable.

\item
If $i_0=n$, choose a model of
\[
\{\neg x_n, x_{n-1},\ldots, x_1\},
\]
which exists by
Theorems~\ref{thm:literal-sat}--\ref{thm:joint-sat}. This model satisfies every clause in 
\[
\{C_{n+1},
  C_{n-1}, \ldots, C_1\}.
\]

Hence, 
\[
S\setminus\{C_n\}
\]
is satisfiable.

\item
If
\[
i_0\in\{1,\ldots,n-1\},
\]
choose a model of

\[
\{\neg x_n,\neg x_{i_0},x_{n-1},\ldots,x_{i_0+1}, x_{i_0-1},\ldots,x_1\},
\]
which exists by
Theorems~\ref{thm:literal-sat}--\ref{thm:joint-sat}. This model satisfies every clause in 
\[
\{C_{n+1}, C_n, C_{n-1}, \ldots, C_{i_0+1},
  C_{i_0-1}, \ldots, C_1\}.
\]

Hence
\[
S\setminus\{C_{i_0}\}
\]
is satisfiable.
\end{enumerate}

Therefore every one-clause deletion from $S$ is satisfiable. Since $S$
itself is unsatisfiable, $S$ is minimally unsatisfiable.

If, starting from the Type-III construction in
Fig.~\ref{fig:delta4-type-III}, one or more admissible interior literals
are added according to the placement rule of
Definition~\ref{def:triangular-standard}, a new triangular
sub-standard contradiction $\Pi(S^*)$ is obtained. We call
$\Pi(S^*)$ a \emph{Type-III$^*$ triangular sub-standard contradiction}.

\begin{remark}
\label{rem:type-II-star}
For each corresponding clause pair $C\subseteq C^*$, every model
satisfying $C$ also satisfies $C^*$. Together with the
minimal-unsatisfiability argument established for admissible extensions,
this implies that the associated clause set $S^*$ remains minimally
unsatisfiable. 
\end{remark}

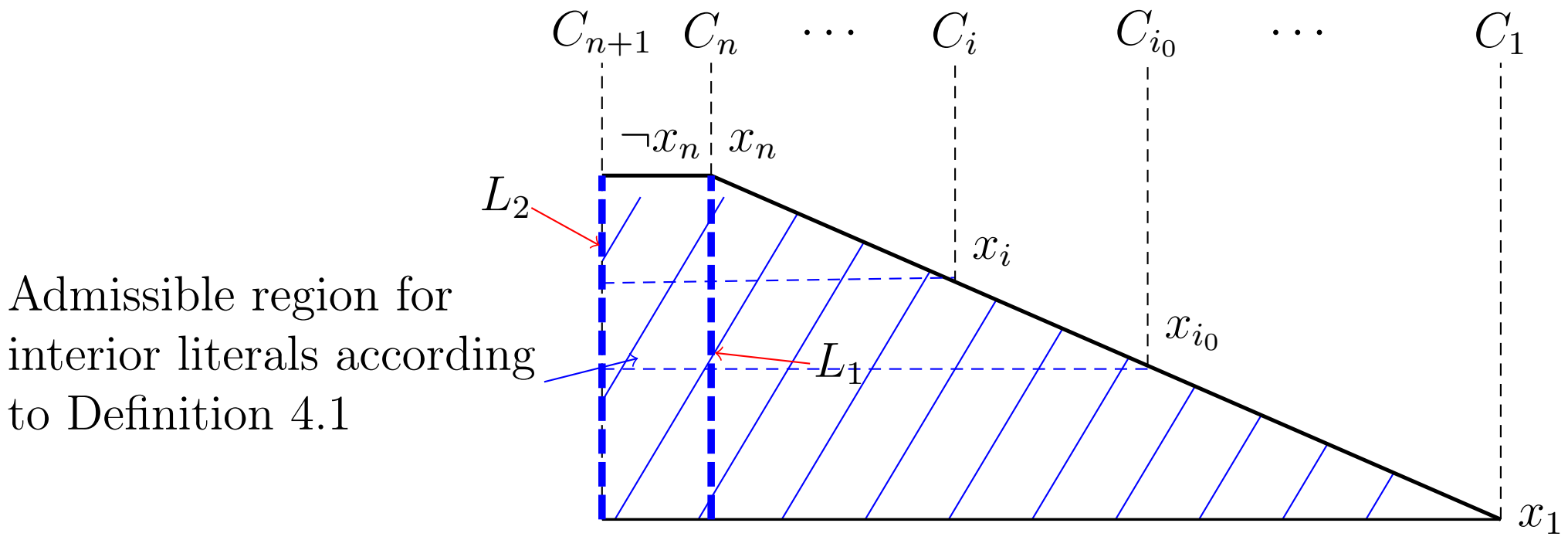
\begin{figure}[t]
\centering
\begin{tikzpicture}[x=1.10cm,y=0.92cm]

% ============================================================
% Key coordinates
% ============================================================
\coordinate (A)  at (0.00,3.20);   % neg x_n
\coordinate (Xn) at (0.85,3.20);   % x_n
\coordinate (Xi) at (2.75,2.20);   % x_i
\coordinate (Xj) at (4.25,1.42);   % x_{i_0}
\coordinate (X1) at (7.00,0.00);   % x_1

% ============================================================
% Hatched admissible interior
% ============================================================
\begin{scope}
  \clip
    (0.00,0.00) --
    (0.00,3.20) --
    (0.85,3.20) --
    (7.00,0.00) --
    cycle;

  \foreach \x in {-1.20,-0.55,...,6.50}
    \draw[blue, thin] (\x,0) -- ++(1.50,3.00);
\end{scope}

% ============================================================
% Main boundary
% ============================================================
\draw[line width=0.95pt]
  (A) -- (Xn) -- (X1);

% middle line one
\draw[blue, densely dashed]
  (0.00,2.20) -- (2.75,2.25);

% middle line two
\draw[blue, densely dashed]
  (0.00,1.40) -- (4.25,1.40);
  
% Bottom boundary
\draw[line width=0.65pt]
  (0.00,0.00) -- (7.00,0.00);

% ============================================================
% Clause lines
% ============================================================

% C_{n+1}
\draw[densely dashed]
  (0.00,0.00) -- (0.00,4.25);

\draw[blue, line width=1.8pt, dash pattern=on 6pt off 2pt]
  (0.00,0.00) -- (0.00,3.20);
  
% C_n
\draw[densely dashed]
(0.85,3.20) -- (0.85,4.25);

\draw[blue, line width=1.8pt, dash pattern=on 6pt off 2pt]
  (0.85,0.00) -- (0.85,3.20);

% C_i
\draw[densely dashed]
  (2.75,2.20) -- (2.75,4.25);

% C_{i_0}
\draw[densely dashed]
  (4.25,1.42) -- (4.25,4.25);

% C_1
\draw[densely dashed]
  (7.00,0.00) -- (7.00,4.25);

% ============================================================
% Clause labels
% ============================================================
\node at (0.00,4.52) {$C_{n+1}$};
\node at (0.85,4.52) {$C_n$};
\node at (1.80,4.52) {$\cdots$};
\node at (2.75,4.52) {$C_i$};
\node at (4.25,4.52) {$C_{i_0}$};
\node at (5.45,4.52) {$\cdots$};
\node at (7.00,4.52) {$C_1$};

% ============================================================
% Main-boundary literals
% ============================================================
\node[above right] at (A)  {$\neg x_n$};
\node[above right] at (Xn) {$x_n$};
\node[above right] at (Xi) {$x_i$};
\node[above right] at (Xj) {$x_{i_0}$};
\node[right]       at (X1) {$x_1$};

% ============================================================
% Representative complementary literals
% ============================================================
%\node[right] at (0.00,2.25) {$\neg x_i$};
%\node[right] at (0.00,1.35) {$\neg x_{i_0}$};

%\node[right] at (0.85,2.25) {$\neg x_j$};
%\node[right] at (0.85,0.20) {$\neg x_1$};

% ============================================================
% L_1 and L_2 annotations
% ============================================================

\node at (1.85,1.45) {$L_1$};
\draw[red, ->]
  (1.62,1.45) -- (0.88,1.55);

\node at (-0.75,3.00) {$L_2$};
\draw[red, ->]
  (-0.55,2.90) -- (-0.02,2.55);

% ============================================================
% Admissible-region annotation
% ============================================================
\node[align=left,anchor=east]
  at (-0.38,1.55)
  {Admissible region for\\
   interior literals according\\
   to Definition~\ref{def:triangular-standard}};

\draw[blue, ->]
  (-0.45,1.28) -- (0.28,1.50);

\end{tikzpicture}

\caption{Schematic representation of a Type-IV triangular
standard contradiction $\Pi(S)$ used in the $\Delta_5$ method.
The sets $L_1$ and $L_2$ represent the two parts of the literals distributed between $C_n$ and $C_{n+1}$, respectively.
The hatched region indicates the admissible region for interior literals
according to Definition~\ref{def:triangular-standard}.}
\label{fig:delta5-type-IV}
\end{figure}

\subsection{$\Delta_5$: Type-IV triangular standard-contradiction method}
\label{sec:delta5}

\noindent As illustrated in Fig.~\ref{fig:delta5-type-IV}, let
\[
\{x_1,\ldots,x_n\}
\]
be a generating literal set whose literals have pairwise distinct
predicate symbols and none of which uses the equality predicate. Let
\[
S=\{C_1,\ldots,C_{n+1}\}.
\]
The literals occurring in $C_{n+1}$ and $C_n$ together form the set
\[
\{\neg x_n,x_n,\neg x_{n-1},\ldots,\neg x_1\},
\]
where
\[
x_n\notin C_{n+1},
\qquad
\neg x_n\notin C_n,
\]
and
\[
C_i=x_i,
\qquad i=1,\ldots,n-1.
\]

The Cartesian-product structure
\[
\Pi(S)=\prod_{i=1}^{n+1}C_i
\]
is a triangular sub-standard contradiction of the corresponding full
triangular standard contradiction. We call $\Pi(S)$ a
\emph{Type-IV triangular standard contradiction}.

Since $\Pi(S)$ is a standard contradiction, the associated clause set
$S$ is unsatisfiable. We next show that deleting any one clause from
$S$ leaves a satisfiable clause set, and hence that $S$ is minimally
unsatisfiable.

Let $C_{i_0}\in S$ (${i_0}=n+1,n,\ldots,1$) be the deleted clause.  According to Theorems~\ref{thm:literal-sat}--\ref{thm:joint-sat}, we have

\begin{enumerate}[label=\arabic*)]

\item
If $i_0=n+1$, choose a model of
\[
\{x_n,\ldots,x_1\},
\]
awhich exists by
Theorems~\ref{thm:literal-sat}--\ref{thm:joint-sat}. This model satisfies every clause in
\[
\{C_n,\ldots,C_1\}.
\]

Hence
\[
S\setminus\{C_{n+1}\}
\]
is satisfiable.

\item
If $i_0=n$, choose a model of
\[
\{\neg x_n,x_{n-1},\ldots,x_1\},
\]
which exists by
Theorems~\ref{thm:literal-sat}--\ref{thm:joint-sat}. This model satisfies every clause in
\[
\{C_{n+1},C_{n-1},\ldots,C_1\}.
\] 

Hence
\[
S\setminus\{C_n\}
\]
is satisfiable.

\item
Suppose
\[
i_0\in\{1,\ldots,n-1\}.
\]
Let $L_1$ ($\not= \varnothing$) and $L_2$ ($\not= \varnothing$) denote the sets of literals occurring,
respectively, in $C_n$ and $C_{n+1}$, so that
\[
L_1\cap L_2=\varnothing,
\qquad
L_1\cup L_2
=
\{\neg x_{n-1},\ldots,\neg x_1\}.
\]

There are two cases.

\begin{enumerate}

\item
If
\[
\neg x_{i_0}\in L_1,
\]
choose a model of
\[
\{\neg x_n,\neg x_{i_0},
x_{n-1},\ldots,x_{i_0+1},
x_{i_0-1},\ldots,x_1\},
\]
which exists by
Theorems~\ref{thm:literal-sat}--\ref{thm:joint-sat}. This model satisfies every clause in 
\[
\{C_{n+1}, C_n, C_{n-1}, \ldots, C_{i_0+1},
  C_{i_0-1}, \ldots, C_1\}.
\]

Hence
\[
S\setminus\{C_{i_0}\}
\]
is satisfiable.

\item
If
\[
\neg x_{i_0}\in L_2,
\]
choose a model of
\[
\{\neg x_{i_0},x_n,
x_{n-1},\ldots,x_{i_0+1},
x_{i_0-1},\ldots,x_1\},
\]
which exists by
Theorems~\ref{thm:literal-sat}--\ref{thm:joint-sat}. This model satisfies every clause in 
\[
\{C_{n+1}, C_n, C_{n-1}, \ldots, C_{i_0+1},
  C_{i_0-1}, \ldots, C_1\}.
\]

Hence
\[
S\setminus\{C_{i_0}\}
\]
is satisfiable.

\end{enumerate}
\end{enumerate}

Thus every one-clause deletion from $S$ is satisfiable. Since $S$
itself is unsatisfiable, $S$ is minimally unsatisfiable.

If one or more admissible interior literals are added to the Type-IV
construction in Fig.~\ref{fig:delta5-type-IV} according to the placement
rule of Definition~\ref{def:triangular-standard}, a new triangular
sub-standard contradiction $\Pi(S^*)$ is obtained. We call $\Pi(S^*)$
a \emph{Type-IV$^*$ triangular sub-standard contradiction}.

\begin{remark}
\label{rem:type-II-star}
For each corresponding clause pair $C\subseteq C^*$, every model
satisfying $C$ also satisfies $C^*$. Together with the
minimal-unsatisfiability argument established for admissible extensions,
this implies that the associated clause set $S^*$ remains minimally
unsatisfiable. 
\end{remark}

\begin{figure}[t]
\centering
\begin{tikzpicture}[x=1.10cm,y=0.92cm]

% ============================================================
% Key coordinates
% ============================================================
\coordinate (A)  at (0.00,3.20);   % neg x_n
\coordinate (Xn) at (0.85,3.20);   % x_n
\coordinate (Xi) at (2.75,2.20);   % x_i
\coordinate (Xj) at (4.25,1.42);   % x_{i_0}
\coordinate (X1) at (7.00,0.00);   % x_1

% ============================================================
% Hatched admissible interior region
% ============================================================
\begin{scope}
  \clip
    (0.00,0.00) --
    (0.00,3.20) --
    (0.85,3.20) --
    (7.00,0.00) --
    cycle;

  \foreach \x in {-1.20,-0.55,...,6.50}
    \draw[blue, thin] (\x,0) -- ++(1.50,3.00);
\end{scope}

% ============================================================
% Main boundary and bottom boundary
% ============================================================
\draw[line width=0.95pt]
  (A) -- (Xn) -- (X1);

% middle line one
\draw[blue, densely dashed]
  (0.00,2.20) -- (2.75,2.25);

% middle line two
\draw[blue, densely dashed]
  (0.00,1.90) -- (3.25,1.90);

% middle line two
\draw[blue, densely dashed]
  (0.00,1.40) -- (4.25,1.40);
  
\draw[line width=0.65pt]
  (0.00,0.00) -- (7.00,0.00);

% ============================================================
% Clause lines
% ============================================================

% C_{n+1}
\draw[densely dashed]
  (0.00,3.20) -- (0.00,4.25);

% C_{n+1}
\draw[line width=0.95pt]
  (0.00,2.20) -- (0.00,3.20);

% C_{n+1}
\draw[densely dashed]
  (0.00,0.00) -- (0.00,2.20);
  
% C_n
\draw[densely dashed]
  (0.85,3.20) -- (0.85,4.25);

% C_i -- extends through the triangular interior
\draw[densely dashed]
  (2.75,2.25) -- (2.75,4.25);

% C_i -- extends through the triangular interior
\draw[line width=0.95pt]
 (2.75,0.00) -- (2.75,1.90);

% C_{i_0}
\draw[densely dashed]
  (4.25,1.42) -- (4.25,4.25);

% C_1
\draw[densely dashed]
  (7.00,0.00) -- (7.00,4.25);

% ============================================================
% Clause labels
% ============================================================
\node at (0.00,4.52) {$C_{n+1}$};
\node at (0.85,4.52) {$C_n$};
\node at (1.80,4.52) {$\cdots$};
\node at (2.75,4.52) {$C_i$};
\node at (4.25,4.52) {$C_{i_0}$};
\node at (5.45,4.52) {$\cdots$};
\node at (7.00,4.52) {$C_1$};

% ============================================================
% Main-boundary literals
% ============================================================
\node[above right] at (A)  {$\neg x_n$};
\node[above right] at (Xn) {$x_n$};
\node[above right] at (Xi) {$x_i$};
\node[above right] at (Xj) {$x_{i_0}$};
\node[right]       at (X1) {$x_1$};

% ============================================================
% Representative complementary literals
% ============================================================

% L_2: literals belonging to C_{n+1}
%\node[right] at (0.00,2.20) {$\neg x_j$};

% L_1: literals belonging to C_i
\node[right] at (3.18,1.95) {$\neg x_{i-1}$};

% ============================================================
% L_2 and L_1 annotations
% ============================================================
\node at (-0.72,3.00) {$L_2$};
\draw[red, ->]
  (-0.52,2.90) -- (-0.02,2.58);

\node at (3.55,1.10) {$L_1$};
\draw[red, ->]
  (3.35,1.10) -- (2.78,1.15);

% ============================================================
% Admissible-region annotation
% ============================================================
\node[align=left,anchor=east]
  at (-0.38,1.55)
  {Admissible region for\\
   interior literals according\\
   to Definition~\ref{def:triangular-standard}};

\draw[blue, ->]
  (-0.25,1.30) -- (0.28,1.52);

\end{tikzpicture}

\caption{Schematic representation of a Type-$V(i)$ triangular
standard contradiction $\Pi(S(i))$ used in the $\Delta_6(i)$ method,
where $i\in\{n-1,\ldots,2\}$. The sets $L_1$ and $L_2$ represent the
two groups of literals separated at the index $i$.
The hatched region indicates the admissible region for interior literals
according to Definition~\ref{def:triangular-standard}.}
\label{fig:delta6-type-V}
\end{figure}

\subsection{$\Delta_6(i)$, $i\in\{n-1,\ldots,2\}$:
Type-$V(i)$ triangular standard-contradiction method}
\label{sec:delta6}

\noindent As illustrated in Fig.~\ref{fig:delta6-type-V}, let
\[
\{x_1,\ldots,x_n\}
\]
be a generating literal set whose literals have pairwise distinct
predicate symbols and none of which uses the equality predicate.
For any
\[
i\in\{n-1,\ldots,2\},
\]
let
\[
S(i)=\{C_{n+1},\ldots,C_1\}
\]
be the clause set associated with the automated theorem generator
$\Delta_6(i)$, where the clauses inside are defined below and their dependence
on $i$ is suppressed for notational simplicity, 

\[
\begin{aligned}
C_{n+1}
  &=\neg x_n\vee\neg x_{n-1}\vee\cdots\vee\neg x_i,\\
C_j
  &=x_j,
  &&j\in\{1,\ldots,n\}\setminus\{i\},\\
C_i
  &=x_i\vee\neg x_{i-1}\vee\cdots\vee\neg x_1.
\end{aligned}
\]
Regarding clauses as sets of literals, we have
\[
\bigl(C_{n+1}\setminus\{\neg x_n\}\bigr)
\cup
\bigl(C_i\setminus\{x_i\}\bigr)
=
\{\neg x_{n-1},\ldots,\neg x_1\}.
\]

The Cartesian-product structure
\[
\Pi(S(i))
=
\prod_{C\in S(i)} C
\]
is a triangular sub-standard contradiction of the corresponding full
triangular standard contradiction. We call $\Pi(S(i))$ a
\emph{Type-$V(i)$ triangular sub-standard contradiction}.

The clause set $S(i)$ is unsatisfiable because $\Pi(S(i))$ is a
standard contradiction. We now show that deleting any one clause from
$S(i)$ leaves a satisfiable clause set. Hence $S(i)$ is minimally
unsatisfiable.

Let $C_{i_0}\in S(i)$ (${i_0}=n+1,n,\ldots,1$) be the deleted clause.  According to Theorems~\ref{thm:literal-sat}--\ref{thm:joint-sat}, we have

\begin{enumerate}[label=\arabic*)]

\item
If $i_0=n+1$, choose a model of
\[
\{x_n,\ldots,x_1\},
\]
which exists by
Theorems~\ref{thm:literal-sat}--\ref{thm:joint-sat}. This model satisfies every clause in
$\{C_n,\ldots,C_1\}$.

Hence
\[
S(i)\setminus\{C_{n+1}\}
\]
is satisfiable.

\item
If $i_0=n$, choose a model of
\[
\{\neg x_n,x_{n-1},\ldots,x_1\},
\]
which exists by
Theorems~\ref{thm:literal-sat}--\ref{thm:joint-sat}. This model satisfies every clause in
\[
\{C_{n+1},C_{n-1},\ldots,C_1\}
\]
is satisfiable. Hence
\[
S(i)\setminus\{C_n\}
\]
is satisfiable.

\item
Suppose
\[
i_0\in\{n-1,\ldots,1\}.
\]
For convenience, define
\[
L_1=C_i\setminus\{x_i\}\not= \varnothing,
\qquad
L_2=C_{n+1}\setminus\{\neg x_n\}\not= \varnothing.
\]
Note that 
\[
L_1\cup L_2
=
\{\neg x_{n-1},\ldots,\neg x_1\},
\qquad
L_1\cap L_2=\varnothing.
\]

There are two cases.

\begin{enumerate}

\item
If
\[
i_0<i,
\]
then
\[
\neg x_{i_0}\in L_1.
\]
choose a model of
\[
\{\neg x_i,\neg x_{i_0}\},
\]
and
\[
\{x_j,
j\in\{1,\ldots,n\}\setminus\{i_0\}\}.
\]
Then $C_{n+1}$ is satisfied by $\neg x_i$,
$C_i$ is satisfied by $\neg x_{i_0}$, and this model satisfies every remaining unit
clause. Hence
\[
S(i)\setminus\{C_{i_0}\}
\]
is satisfiable.

\item
If
\[
i_0\geq i,
\]
then
\[
\neg x_{i_0}\in L_2.
\]
choose a model of
\[
\neg x_{i_0}
\]
and 
\[
\{x_j, j\in\{1,\ldots,n\}\setminus\{i_0\}\}.
\]
Then $C_{n+1}$ is satisfied by $\neg x_{i_0}$,
$C_i$ is satisfied by $x_i$ whenever $i_0\not=i$, while if $i_0=i$
the clause $C_i$ has been deleted, and this model satisfies all remaining unit clauses. Hence
\[
S(i)\setminus\{C_{i_0}\}
\]
is satisfiable.

\end{enumerate}
\end{enumerate}

Thus every one-clause deletion from $S(i)$ is satisfiable. Since
$S(i)$ itself is unsatisfiable, $S(i)$ is minimally unsatisfiable.

For any $i\in\{n-1,\ldots,2\}$, if one or more admissible interior
literals are added to the Type-$V(i)$ construction in
Fig.~\ref{fig:delta6-type-V} according to the placement rule of
Definition~\ref{def:triangular-standard}, a new triangular
sub-standard contradiction $\Pi(S(i)^*)$ is obtained. We call it a
\emph{Type-$V(i)^*$ triangular sub-standard contradiction} generated
from $\Pi(S(i))$.

\begin{remark}
For every corresponding pair of clauses
\[
C\in S(i),
\qquad
C^*\in S(i)^*,
\]
we have
\[
C\subseteq C^*.
\]
By the minimal-unsatisfiability preservation result for admissible
extensions, the associated clause set $S(i)^*$ remains minimally
unsatisfiable. 
\end{remark}

\begin{figure}[t]
\centering
\begin{tikzpicture}[x=1.10cm,y=0.92cm]

% ============================================================
% Key coordinates
% ============================================================
\coordinate (A)  at (0.00,3.20);   % neg x_n
\coordinate (Xn) at (0.85,3.20);   % x_n
\coordinate (Xi) at (2.75,2.20);   % x_i
\coordinate (Xj) at (4.25,1.42);   % x_{i_0}
\coordinate (X1) at (7.00,0.00);   % x_1

% ============================================================
% Hatched admissible interior region
% ============================================================
\begin{scope}
  \clip
    (0.00,0.00) --
    (0.00,3.20) --
    (0.85,3.20) --
    (7.00,0.00) --
    cycle;

  \foreach \x in {-1.20,-0.55,...,6.50}
    \draw[blue, thin] (\x,0) -- ++(1.50,3.00);
\end{scope}

% ============================================================
% Main boundary and bottom boundary
% ============================================================
\draw[line width=0.95pt]
  (A) -- (Xn) -- (X1);

% middle line one
\draw[blue, densely dashed]
  (0.00,2.20) -- (2.75,2.25);

% middle line two
\draw[blue, densely dashed]
  (0.00,1.90) -- (3.25,1.90);

% middle line two
\draw[blue, densely dashed]
  (0.00,1.40) -- (4.25,1.40);
  
\draw[line width=0.65pt]
  (0.00,0.00) -- (7.00,0.00);

% ============================================================
% Clause lines
% ============================================================

% C_{n+1}
\draw[densely dashed]
(0.00,0.00) -- (0.00,4.25);

% C_n: extends into the triangular interior
\draw[densely dashed]
  (0.85,3.20) -- (0.85,4.25);

% C_n
\draw[line width=0.95pt]
  (0.85,2.20) -- (0.85,3.20);

% C_n
%\draw[densely dashed]
%  (0.00,0.00) -- (0.00,2.20);
  
% C_i -- extends through the triangular interior
\draw[densely dashed]
  (2.75,2.25) -- (2.75,4.25);

% C_i -- extends through the triangular interior
\draw[line width=0.95pt]
 (2.75,0.00) -- (2.75,1.90);

% C_{i_0}
\draw[densely dashed]
  (4.25,1.42) -- (4.25,4.25);

% C_1
\draw[densely dashed]
  (7.00,0.00) -- (7.00,4.25);

% ==============================

==============================
% Horizontal dotted guides
% ============================================================
%\draw[densely dotted]
%  (0.00,2.45) -- (2.75,2.45);

%\draw[densely dotted]
%  (0.00,1.65) -- (4.25,1.65);

%\draw[densely dotted]
%  (0.00,0.85) -- (5.70,0.85);

% ============================================================
% Clause labels
% ============================================================
\node at (0.00,4.52) {$C_{n+1}$};
\node at (0.85,4.52) {$C_n$};
\node at (1.80,4.52) {$\cdots$};
\node at (2.75,4.52) {$C_i$};
\node at (4.25,4.52) {$C_{i_0}$};
\node at (5.45,4.52) {$\cdots$};
\node at (7.00,4.52) {$C_1$};

% ============================================================
% Main-boundary literals
% ============================================================
\node[above right] at (A)  {$\neg x_n$};
\node[above right] at (Xn) {$x_n$};
\node[above right] at (Xi) {$x_i$};
\node[above right] at (Xj) {$x_{i_0}$};
\node[right]       at (X1) {$x_1$};

% Representative endpoint literal at the split
\node[above right] at (3.15,1.8) {$x_{i-1}$};

% ============================================================
% L_2 and L_1 annotations
% ============================================================

% L_2 = C_n \ {x_n}
\node at (-0.72,3.02) {$L_2$};
\draw[red, ->]
  (-0.48,2.92) -- (0.82,2.58);

% L_1 = C_i \ {x_i}
\node at (3.72,1.10) {$L_1$};
\draw[red, ->]
  (3.48,1.12) -- (2.78,1.25);

% ============================================================
% Admissible-region annotation
% ============================================================
\node[align=left,anchor=east]
  at (-0.38,1.55)
  {Admissible region for\\
   interior literals according\\
   to Definition~\ref{def:triangular-standard}};

\draw[blue, ->]
  (-0.4,1.30) -- (0.28,1.52);

\end{tikzpicture}

\caption{Schematic representation of a Type-$VI(i)$ triangular
standard contradiction $\Pi(S(i))$ used in the $\Delta_7(i)$ method,
where $i\in\{n-1,\ldots,2\}$. The sets
$L_1=C_i\setminus\{x_i\}$ and
$L_2=C_n\setminus\{x_n\}$ represent the two groups of complementary
literals. The hatched region indicates the admissible region for
interior literals according to
Definition~\ref{def:triangular-standard}.}
\label{fig:delta7-type-VI}
\end{figure}
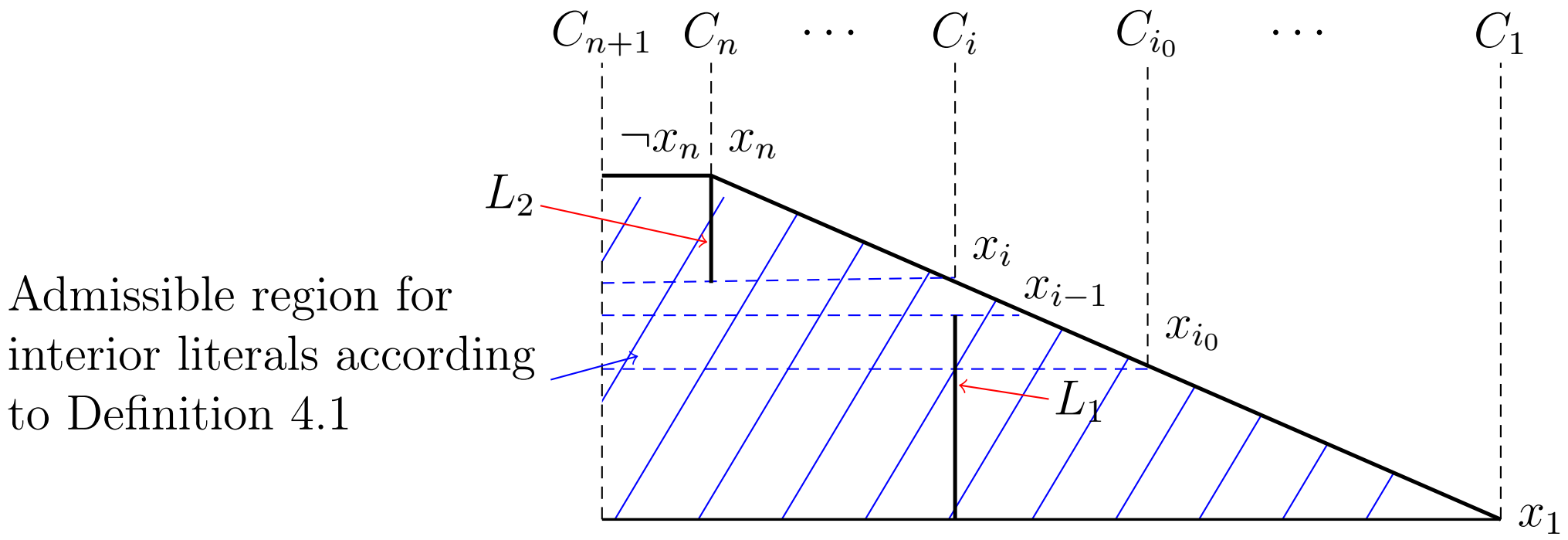

\subsection{$\Delta_7(i)$, $i\in\{n-1,\ldots,2\}$:
Type-$VI(i)$ triangular standard-contradiction method}
\label{sec:delta7}

\noindent As illustrated in Fig.~\ref{fig:delta7-type-VI}, let
\[
\{x_1,\ldots,x_n\}
\]
be a generating literal set whose literals have pairwise distinct
predicate symbols and none of which uses the equality predicate.
For any
\[
i\in\{n-1,\ldots,2\},
\]
let
\[
S(i)=\{C_1,\ldots,C_{n+1}\}
\]
be the clause set associated with the automated theorem generator
$\Delta_7(i)$, where
\[
C_{n+1}=\neg x_n,
\]
\[
C_n
=
x_n\vee\neg x_{n-1}\vee\cdots\vee\neg x_i,
\]
\[
C_i
=
x_i\vee\neg x_{i-1}\vee\cdots\vee\neg x_1,
\]
and
\[
C_j=x_j,
\qquad
j\in\{1,\ldots,n-1\}\setminus\{i\}.
\]

Identifying each clause with its set of literals, we have
\[
\bigl(C_n\setminus\{x_n\}\bigr)
\cup
\bigl(C_i\setminus\{x_i\}\bigr)
=
\{\neg x_{n-1},\ldots,\neg x_1\}.
\]
Moreover, the two sets on the left-hand side are disjoint.

The Cartesian-product structure
\[
\Pi(S(i))
=
\prod_{C\in S(i)} C
\]
is a triangular sub-standard contradiction of the corresponding full
triangular standard contradiction. We call $\Pi(S(i))$ a
\emph{Type-$VI(i)$ triangular sub-standard contradiction}.

The clause set $S(i)$ is unsatisfiable because $\Pi(S(i))$ is a
standard contradiction. We now show that deleting any one clause from
$S(i)$ leaves a satisfiable clause set. Hence $S(i)$ is minimally
unsatisfiable.

Let $C_{i_0}\in S(i)$ (${i_0}=n+1,n,\ldots,1$) be the deleted clause.  According to Theorems~\ref{thm:literal-sat}--\ref{thm:joint-sat}, we have

\begin{enumerate}[label=\arabic*)]

\item
If $i_0=n+1$, choose a model of
\[
\{x_n,\ldots,x_1\},
\]
which exists by
Theorems~\ref{thm:literal-sat}--\ref{thm:joint-sat}. This model satisfies every clause in
$\{C_n,\ldots,C_1\}$. Hence
\[
S(i)\setminus\{C_{n+1}\}
\]
is satisfiable.

\item
If $i_0=n$, choose a model of
\[
\{\neg x_n,x_{n-1},\ldots,x_1\},
\]
which exists by
Theorems~\ref{thm:literal-sat}--\ref{thm:joint-sat}. This model satisfies every clause in
\[
\{C_{n+1},C_{n-1},\ldots,C_1\}.
\]
Hence
\[
S(i)\setminus\{C_n\}
\]
is satisfiable.

\item
Suppose
\[
i_0\in\{n-1,\ldots,1\}.
\]
For convenience, let
\[
L_1=C_i\setminus\{x_i\}\not= \varnothing,
\qquad
L_2=C_n\setminus\{x_n\}\not= \varnothing.
\]
Note that
\[
L_1\cup L_2
=
\{\neg x_{n-1},\ldots,\neg x_1\},
\qquad
L_1\cap L_2=\varnothing.
\]
\end{enumerate}
There are two cases.

\begin{enumerate}

\item
If $i_0<i$, then
\[
\neg x_{i_0}\in L_1.
\]
Choose a model of
\[
\{\neg x_n, \neg x_{i}, \neg x_{i_0}\},
\]
and
\[
\{x_j, j\in\{1,\ldots,n-1\}\setminus\{i_0, i\}\}, 
\]
which exists by
Theorems~\ref{thm:literal-sat}--\ref{thm:joint-sat}. Then $C_{n+1}$ is satisfied by $\neg x_n$, $C_{n}$ is satisfied by $\neg x_i$, 
$C_i$ is satisfied by $\neg x_{i_0}$, and this model satisfies all remaining unit clauses. Hence
\[
S(i)\setminus\{C_{i_0}\}
\]
is satisfiable.

\item
If $i_0\geq i$, then
\[
\neg x_{i_0}\in L_2.
\]
Choose a model of
\[
\{\neg x_n,\neg x_{i_0}\},
\]
and
\[
\{x_j, j\in\{1,\ldots,n-1\}\setminus\{i_0, i\}\},
\]
which exists by
Theorems~\ref{thm:literal-sat}--\ref{thm:joint-sat}. Then $C_{n+1}$ is satisfied by $\neg x_n$,
$C_n$ is satisfied by $\neg x_{i_0}$, $C_i$ is satisfied by $x_i$,
and this model satisfies all remaining unit clauses. Hence
\[
S(i)\setminus\{C_{i_0}\}
\]
is satisfiable.

\end{enumerate}

Thus every one-clause deletion from $S(i)$ is satisfiable. Since
$S(i)$ itself is unsatisfiable, $S(i)$ is minimally unsatisfiable.

For any $i\in\{n-1,\ldots,2\}$, if one or more admissible interior
literals are added to the Type-$VI(i)$ construction in
Fig.~\ref{fig:delta7-type-VI} according to the placement rule of
Definition~\ref{def:triangular-standard}, a new triangular
sub-standard contradiction $\Pi(S(i)^*)$ is obtained. We call
$\Pi(S(i)^*)$ a \emph{Type-$VI(i)^*$ triangular sub-standard
contradiction} generated from $\Pi(S(i))$.

\begin{remark}
For each corresponding pair
\[
C\in S(i),
\qquad
C^*\in S(i)^*,
\]
we have
\[
C\subseteq C^*.
\]
By the minimal-unsatisfiability preservation result for admissible
extensions, the associated clause set $S(i)^*$ remains minimally
unsatisfiable. 
\end{remark}

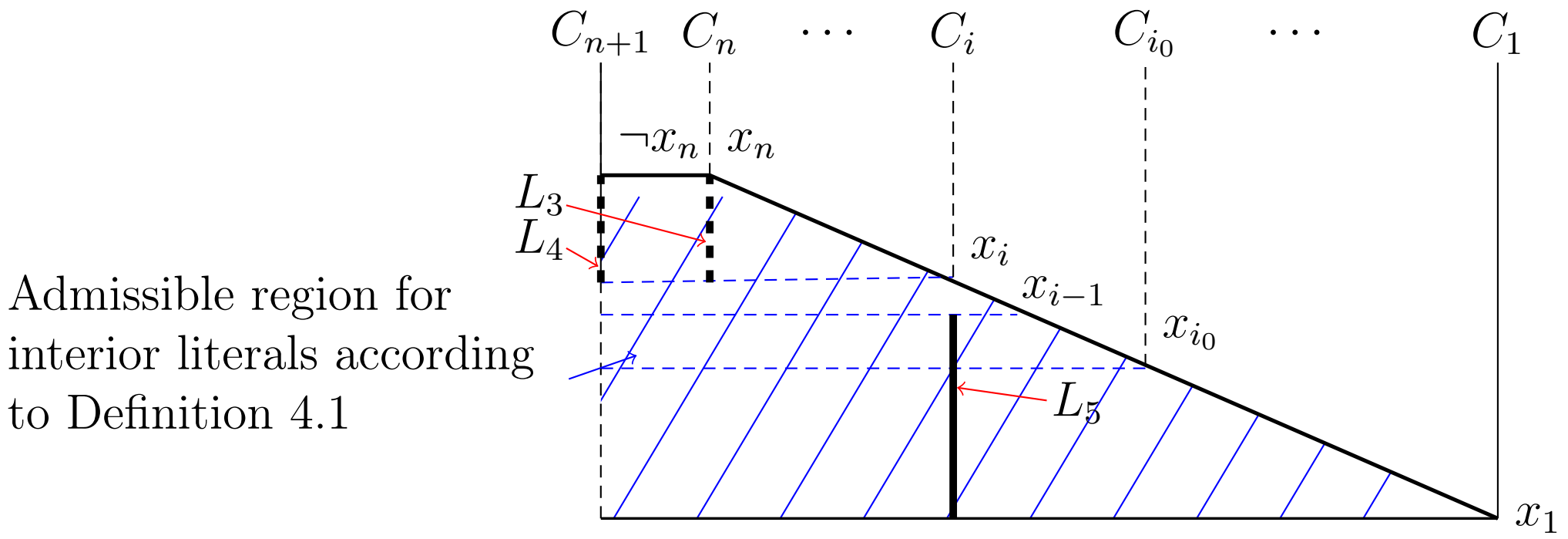
\begin{figure}[t]
\centering
\begin{tikzpicture}[x=1.10cm,y=0.92cm]

% ============================================================
% Key coordinates
% ============================================================
\coordinate (A)  at (0.00,3.20);   % neg x_n
\coordinate (Xn) at (0.85,3.20);   % x_n
\coordinate (Xi) at (2.75,2.20);   % x_i
\coordinate (Xj) at (4.25,1.42);   % x_{i_0}
\coordinate (X1) at (7.00,0.00);   % x_1

% ============================================================
% Hatched admissible interior region
% ============================================================
\begin{scope}
  \clip
    (0.00,0.00) --
    (0.00,3.20) --
    (0.85,3.20) --
    (7.00,0.00) --
    cycle;

  \foreach \x in {-1.20,-0.55,...,6.50}
    \draw[blue, thin] (\x,0) -- ++(1.50,3.00);
\end{scope}

% ============================================================
% Main boundary and bottom boundary
% ============================================================
\draw[line width=0.95pt]
  (A) -- (Xn) -- (X1);

% middle line one
\draw[blue, densely dashed]
  (0.00,2.20) -- (2.75,2.25);

% middle line two
\draw[blue, densely dashed]
  (0.00,1.90) -- (3.25,1.90);

% middle line two
\draw[blue, densely dashed]
  (0.00,1.40) -- (4.25,1.40);

\draw[line width=0.65pt]
  (0.00,0.00) -- (7.00,0.00);

% ============================================================
% Clause lines
% ============================================================

% C_{n+1}: neg x_n together with L_4
\draw
  (0.00,2.20) -- (0.00,4.25);

% C_{n+1}
\draw[densely dashed]
  (0.00,3.20) -- (0.00,4.25);

% C_{n+1}
\draw[line width=1.85pt, dashed]
  (0.00,2.20) -- (0.00,3.20);

% C_{n+1}
\draw[densely dashed]
  (0.00,0.00) -- (0.00,2.20);

% C_n: extends into the triangular interior
\draw[densely dashed]
  (0.85,3.20) -- (0.85,4.25);

% C_n
\draw[line width=1.85pt, dashed]
  (0.85,2.20) -- (0.85,3.20);

% C_i -- extends through the triangular interior
\draw[densely dashed]
  (2.75,2.25) -- (2.75,4.25);

% C_i -- extends through the triangular interior
\draw[line width=1.85pt]
 (2.75,0.00) -- (2.75,1.90);

% C_{i_0}
\draw[densely dashed]
  (4.25,1.42) -- (4.25,4.25);

% C_1
\draw
  (7.00,0.00) -- (7.00,4.25);

% ============================================================
% Clause labels
% ============================================================
\node at (0.00,4.52) {$C_{n+1}$};
\node at (0.85,4.52) {$C_n$};
\node at (1.80,4.52) {$\cdots$};
\node at (2.75,4.52) {$C_i$};
\node at (4.25,4.52) {$C_{i_0}$};
\node at (5.45,4.52) {$\cdots$};
\node at (7.00,4.52) {$C_1$};

% ============================================================
% Main-boundary literals
% ============================================================
\node[above right] at (A)  {$\neg x_n$};
\node[above right] at (Xn) {$x_n$};
\node[above right] at (Xi) {$x_i$};
\node[above right] at (Xj) {$x_{i_0}$};
\node[right]       at (X1) {$x_1$};

% Literal immediately following x_i on the main boundary
\node[above right] at (3.15,1.8) {$x_{i-1}$};

% ============================================================
% L_3, L_4, L_5
% ============================================================

% L_3 associated with C_n
\node at (-0.48,3.02) {$L_3$};
\draw[red, ->]
  (-0.27,2.92) -- (0.82,2.58);

% L_4 associated with C_{n+1}
\node at (-0.48,2.60) {$L_4$};
\draw[red, ->]
  (-0.27,2.52) -- (-0.02,2.35);

% L_5 associated with C_i
\node at (3.72,1.08) {$L_5$};
\draw[red, ->]
  (3.48,1.10) -- (2.78,1.22);

% ============================================================
% Admissible-region annotation
% ============================================================
\node[align=left,anchor=east]
  at (-0.38,1.55)
  {Admissible region for\\
   interior literals according\\
   to Definition~\ref{def:triangular-standard}};

\draw[blue, ->]
  (-0.25,1.30) -- (0.28,1.52);

\end{tikzpicture}

\caption{Schematic representation of a Type-$VII(i)$ triangular
standard contradiction $\Pi(S(i))$ used in the $\Delta_8(i)$
method, where $i\in\{n-1,\ldots,2\}$. The sets $L_3$, $L_4$, and
$L_5$ represent the groups of literals distributed
among $C_n$, $C_{n+1}$, and $C_i$, respectively. The hatched region
indicates the admissible region for interior literals according to
Definition~\ref{def:triangular-standard}.}
\label{fig:delta8-type-VII}
\end{figure}

\subsection{$\Delta_8(i)$, $i\in\{n-1,\ldots,2\}$:
Type-$VII(i)$ triangular sub-standard-contradiction method}
\label{sec:delta8}

As illustrated in Fig.~\ref{fig:delta8-type-VII}, let
$\{x_1,\ldots,x_n\}$ be a generating literal set whose
members have pairwise distinct predicate symbols and
do not involve equality.

For any $i\in\{n-1,\ldots,2\}$, let
\[
S(i)=\{C_{n+1},\ldots,C_1\},
\]
which corresponds to the automated theorem generator
$\Dlt_8(i)$. The dependence of the clauses inside on
the parameter $i$ is suppressed for notational
simplicity.

For convenience, the clauses are represented as sets
of literals, where
\[
\begin{aligned}
C_{n+1} &= \{\neg x_n\}\cup L_4,\\
C_n     &= \{x_n\}\cup L_3,\\
C_j     &= \{x_j\},
\quad j\in\{1,\ldots,n-1\}\setminus\{i\},\\
C_i     &= \{x_i\}\cup
           \{\neg x_{i-1},\ldots,\neg x_1\}.
\end{aligned}
\]
Regarding clauses as sets of literals, we have
\[
\begin{aligned}
&\bigl(C_{n+1}\setminus\{\neg x_n\}\bigr)
\cup
\bigl(C_n\setminus\{x_n\}\bigr)\\
&\qquad\cup
\bigl(C_i\setminus\{x_i\}\bigr)
=
\{\neg x_{n-1},\ldots,\neg x_1\}.
\end{aligned}
\]

Here, let
\[
\begin{aligned}
L_5 &= C_i\setminus\{x_i\},\\
L_4 &= C_{n+1}\setminus\{\neg x_n\},\\
L_3 &= C_n\setminus\{x_n\}.
\end{aligned}
\]

When $n-1=i$,
\[
L_5=\{\neg x_{n-2},\ldots,\neg x_1\},
\]
and either
\[
L_4=\{\neg x_{n-1}\},
\qquad L_3=\varnothing,
\]
or
\[
L_4=\varnothing,
\qquad L_3=\{\neg x_{n-1}\}.
\]

When $n-1>i$,
\[
\begin{aligned}
L_5 &= \{\neg x_{i-1},\ldots,\neg x_1\},\\
L_4\cup L_3
    &= \{\neg x_{n-1},\ldots,\neg x_i\},\\
L_4\cap L_3 &= \varnothing,\\
L_4 &\neq\varnothing,\\
L_3 &\neq\varnothing.
\end{aligned}
\]

The associated Cartesian-product structure $\Pi(S(i))$
is a triangular sub-standard contradiction of a full
triangular standard contradiction and is therefore
unsatisfiable. Such triangular sub-standard contradictions
are called \emph{Type-$VII(i)$ triangular standard
contradictions}.

Since $\Pi(S(i))$ is a standard contradiction, the associated clause
set $S(i)$ is unsatisfiable. We now show that deleting any one clause
from $S(i)$ leaves a satisfiable clause set. Hence $S(i)$ is minimally
unsatisfiable.

Let $C_{i_0}\in S(i)$ (${i_0}=n+1,n,\ldots,1$) be the deleted clause.  According to Theorems~\ref{thm:literal-sat}--\ref{thm:joint-sat}, we have

\begin{enumerate}

\item
If $i_0=n+1$, choose a model of
\[
\{x_n,\ldots,x_1\},
\]
which exists by
Theorems~\ref{thm:literal-sat}--\ref{thm:joint-sat}. This model satisfies every clause in
\[
\{C_n,\ldots,C_1\}.
\]
Hence
\[
S(i)\setminus\{C_{n+1}\}
\]
is satisfiable.

\item
If $i_0=n$, choose a model of
\[
\{\neg x_n,x_{n-1},\ldots,x_1\},
\]
which exists by
Theorems~\ref{thm:literal-sat}--\ref{thm:joint-sat}. This model satisfies every clause in
\[
\{C_{n+1},C_{n-1},\ldots,C_1\}.
\]
Hence
\[
S(i)\setminus\{C_n\}
\]
is satisfiable.

\item
Suppose that
$i_0\in\{n-1,\ldots,1\}$.
Note that
\[
L_5\cup L_4\cup L_3
=
\{\neg x_{n-1},\ldots,\neg x_1\}.
\]

\medskip
\noindent
{(i) When $i_0<i$:}

In this case, $\neg x_{i_0}\in L_5$.

\smallskip
\noindent
If $\neg x_i\in L_3$, choose a model of
\[
\{\neg x_n,\neg x_i,
x_{n-1},\ldots,x_{i+1},
\neg x_{i_0},
x_{i-1},\ldots,x_{i_0+1},
x_{i_0-1},\ldots,x_1\},
\]
which exists by
Theorems~\ref{thm:literal-sat}--\ref{thm:joint-sat}. This model satisfies every clause in
\[
\{C_{n+1},C_n,C_{n-1},\ldots,C_{i+1},
C_i,C_{i-1},\ldots,C_{i_0+1},
C_{i_0-1},\ldots,C_1\}.
\]
Hence,
\[
S(i)\setminus\{C_{i_0}\}
\]
is satisfiable.

\smallskip
\noindent
If $\neg x_i\in L_4$, choose a model of
\[
\{\neg x_i,x_n,
x_{n-1},\ldots,x_{i+1},
\neg x_{i_0},
x_{i-1},\ldots,x_{i_0+1},
x_{i_0-1},\ldots,x_1\},
\]
which exists by
Theorems~\ref{thm:literal-sat}--\ref{thm:joint-sat}. This model satisfies every clause in
\[
\{C_{n+1},C_n,C_{n-1},\ldots,C_{i+1},
C_i,C_{i-1},\ldots,C_{i_0+1},
C_{i_0-1},\ldots,C_1\}.
\]
Hence,
\[
S(i)\setminus\{C_{i_0}\}
\]
is satisfiable.

\medskip
\noindent
(ii) When $i_0\ge i$:

\smallskip
\noindent
If $\neg x_{i_0}\in L_4$, choose a model of
\[
\{\neg x_{i_0},x_n,
x_{n-1},\ldots,x_{i_0+1},
x_{i_0-1},\ldots,x_1\},
\]
which exists by
Theorems~\ref{thm:literal-sat}--\ref{thm:joint-sat}. This model satisfies every clause in
\[
\{C_{n+1},C_n,C_{n-1},\ldots,C_{i_0+1},
C_{i_0-1},\ldots,C_1\}.
\]
Hence,
\[
S(i)\setminus\{C_{i_0}\}
\]
is satisfiable.

\smallskip
\noindent
If $\neg x_{i_0}\in L_3$, choose a model of
\[
\{\neg x_n,\neg x_{i_0},
x_{n-1},\ldots,x_{i_0+1},
x_{i_0-1},\ldots,x_1\},
\]
which exists by
Theorems~\ref{thm:literal-sat}--\ref{thm:joint-sat}. This model satisfies every clause in
\[
\{C_{n+1},C_n,C_{n-1},\ldots,C_{i_0+1},
C_{i_0-1},\ldots,C_1\}.
\]
Hence,
\[
S(i)\setminus\{C_{i_0}\}
\]
is satisfiable.

\end{enumerate}

Thus every one-clause deletion from $S(i)$ is satisfiable. Since
$S(i)$ itself is unsatisfiable, $S(i)$ is minimally unsatisfiable.

For any $i\in\{n-1,\ldots,2\}$, if one or more admissible interior
literals are added to the Type-$VII(i)$ construction in
Fig.~\ref{fig:delta8-type-VII} according to the placement rule of
Definition~\ref{def:triangular-standard}, a new triangular
sub-standard contradiction $\Pi(S(i)^*)$ is obtained. We call it a
\emph{Type-$VII(i)^*$ triangular sub-standard contradiction} generated
from $\Pi(S(i))$.

For every corresponding pair
\[
C\in S(i),
\qquad
C^*\in S(i)^*,
\]
we have
\[
C\subseteq C^*.
\]
By the minimal-unsatisfiability preservation result for admissible
extensions, the associated clause set $S(i)^*$ remains minimally
unsatisfiable. 

The preceding $\Dlt_1,\Dlt_2,\Dlt_3,\Dlt_4,\Dlt_5,\Dlt_6(i),\Dlt_7(i),\Dlt_8(i)$ constructions give eight specific families of methods for constructing minimally unsatisfiable clause sets. We now give the ninth, general $\Dlt$-class construction.

\subsection{A general method for generating minimally unsatisfiable clause sets from $\Dlt$-class triangular standard contradictions}

\begin{lemma}
\label{lem:clause-deletion}
Let $S$ be a first-order clause set, and let $C\in S$.
Suppose that $y$ is a literal occurring in $C$ whose
predicate symbol is distinct from the predicate symbols
of all other literals in $S$.

Then $S$ and
\[
S'=S\setminus\{C\}
\]
are equisatisfiable. That is, 
\[
S \text{ is satisfiable}
\quad\Longleftrightarrow\quad
S\setminus\{C\} \text{ is satisfiable}.
\]
\end{lemma}
\begin{proof}
See~\ref{app:lem51}.
\end{proof}

\begin{lemma}\label{lem:add}
Let $S$ be a satisfiable first-order clause set and let $y$ be a unit clause whose predicate symbol is different from every predicate symbol occurring in $S$. Then $S'=S\cup\{y\}$ is satisfiable.
\end{lemma}
\begin{proof}
See~\ref{app:lem52}.
\end{proof}

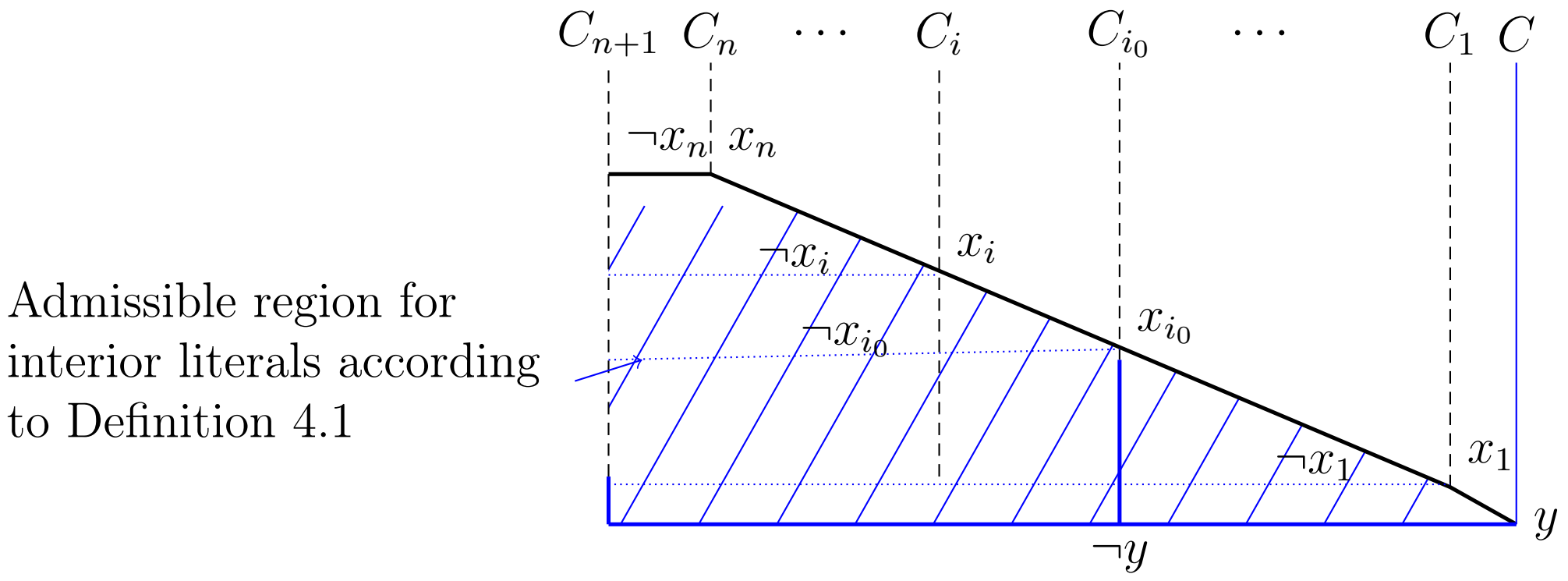
\begin{figure}[t]
\centering
\begin{tikzpicture}[x=1.02cm,y=0.90cm]

% ============================================================
% Main coordinates
% ============================================================
\coordinate (A)   at (0.00,3.30);  % neg x_n
\coordinate (Xn)  at (0.85,3.30);  % x_n
\coordinate (Xi)  at (2.75,2.30);  % x_i
\coordinate (Xj)  at (4.25,1.55);  % x_{i_0}
\coordinate (X1)  at (7.00,0.35);  % x_1
\coordinate (Y)   at (7.55,0.00);  % y

% ============================================================
% Hatched admissible interior
% ============================================================
\begin{scope}
  \clip
    (0.00,0.00) --
    (0.00,3.30) --
    (0.85,3.30) --
    (7.00,0.35) --
    (7.55,0.00) --
    cycle;

  \foreach \x in {-1.2,-0.55,...,7.0}
    \draw[blue, thin] (\x,0) -- ++(1.50,3.00);
\end{scope}

% ============================================================
% Main boundary
% ============================================================
\draw[line width=0.95pt]
  (A) -- (Xn) -- (X1) -- (Y);

% Bottom boundary
\draw[blue, line width=0.95pt]
  (0.00,0.00) -- (Y);

% ============================================================
% Clause lines
% ============================================================
\draw[densely dashed]
(0.00,0.45) -- (0.00,4.35);        % C_{n+1}

\draw[blue, line width=0.95pt]
(0.00,0.00) -- (0.00,0.45);        % C_{n+1}

\draw[densely dashed] 
(0.85,3.30) -- (0.85,4.35);        % C_n

\draw[densely dashed]
  (2.75,0.45) -- (2.75,4.35);            % C_i

\draw[densely dashed]
  (4.25,1.55) -- (4.25,4.35);            % C_{i_0}
  
\draw[blue, line width=0.95pt]
  (4.25,0.00) -- (4.25,1.55);            % C_{i_0}

\draw[densely dashed] 
(7.00,0.35) -- (7.00,4.35);        % C_1

% New clause C
\draw [blue]
(7.55,0.00) -- (7.55,4.35);

% ============================================================
% Horizontal placement guides
% ============================================================
\draw[blue, densely dotted] (0.00,2.35) -- (2.75,2.35);
\draw[blue, densely dotted] (0.00,1.55) -- (4.25,1.65);
\draw[blue, yshift=-7pt, densely dotted] (0.00,0.65) -- (7.00,0.65);

% ============================================================
% Clause labels
% ============================================================
\node at (0.00,4.62) {$C_{n+1}$};
\node at (0.85,4.62) {$C_n$};
\node at (1.80,4.62) {$\cdots$};
\node at (2.75,4.62) {$C_i$};
\node at (4.25,4.62) {$C_{i_0}$};
\node at (5.45,4.62) {$\cdots$};
\node at (7.00,4.62) {$C_1$};
\node at (7.55,4.62) {$C$};

% ============================================================
% Literal labels
% ============================================================
\node[above right] at (A)  {$\neg x_n$};
\node[above right] at (Xn) {$x_n$};

\node[above left]  at (2.00,2.20) {$\neg x_i$};
\node[above right] at (Xi) {$x_i$};

\node[above left]  at (2.5,1.42) {$\neg x_{i_0}$};
\node[above right] at (Xj) {$x_{i_0}$};

\node[above left] at (6.35,0.25) {$\neg x_1$};
\node[above right] at (X1) {$x_1$};

% New complementary pair
\node[below] at (4.25,0.00) {$\neg y$};
\node[right] at (Y) {$y$};

% ============================================================
% Annotation
% ============================================================
\node[align=left,anchor=east]
  at (-0.42,1.55)
  {Admissible region for\\
   interior literals according\\
   to Definition~\ref{def:triangular-standard}};

\draw[blue, ->]
  (-0.28,1.35) -- (0.28,1.55);

\end{tikzpicture}

\caption{Schematic representation of the general construction of a
minimally unsatisfiable clause set from a triangular standard
contradiction. The literal $y$ has a predicate symbol that occurs
nowhere else in the clause set.}
\label{fig:general-triangular-mu}
\end{figure}

\begin{theorem}[General $\Dlt$-class construction]\label{thm:general}
Let $\{x_1,\ldots,x_n\}$ be the generating literal set of a triangular standard contradiction in first-order logic, with pairwise distinct predicate symbols and no equality predicate.
\begin{enumerate}[label=\Roman*.]
\item If clause $C_i$ contains $x_i$, the row containing $x_i$ contains exactly one occurrence of $\neg x_i$ for each $i=1,\ldots,n$, and $C_{n+1}=\neg x_n$ (as schematically shown in Fig.~\ref{fig:general-triangular-mu}), then $S=\{C_{n+1},C_n,\ldots,C_1\}$ is a minimally unsatisfiable clause set.
\item Such a $\Dlt$-class triangular standard contradiction is also a minimal-unsatisfiable construction with the minimum number of literals, namely $2n$ literals in total.
\end{enumerate}
\end{theorem}
\begin{proof}
See~\ref{app:thm51}.
\end{proof}

\begin{remark}
The general method for generating minimally unsatisfiable clause sets
from $\Delta$-type triangular standard contradictions does not
encompass the $\Dlt_1$ method.
\end{remark}

\begin{remark}
\label{rem:general-extension}
Let $\Pi(S)$ be a triangular standard contradiction generated
from the generating literal set $\{x_1,\ldots,x_n\}$, where
$S$ is the minimally unsatisfiable clause set specified in
Theorem~\ref{thm:general}.

Deleting any clause $C_{i_0}$ from $S$, where
$i_0\in\{1,\ldots,n+1\}$, leaves a satisfiable clause set
$S\setminus\{C_{i_0}\}$.

If an arbitrary nonempty set of literals is placed in the
triangular interior (the region indicated by the oblique
dashed line in Fig.~\ref{fig:general-triangular-mu}), according to the
literal-placement principle of Definition~\ref{def:triangular-standard}, the resulting
Type-$\Delta$ triangular standard contradiction has an
associated clause set $S^*$.

By Theorems~\ref{thm:literal-sat}
and~\ref{thm:joint-sat}, deleting the clause
$C_{i_0}^*$ corresponding to $C_{i_0}$ from $S^*$
also leaves a satisfiable clause set
\[
S^*\setminus\{C_{i_0}^*\}.
\]

Indeed, for every clause $C\in S\setminus\{C_{i_0}\}$
and its corresponding clause
$C^*\in S^*\setminus\{C_{i_0}^*\}$, regarding clauses
as sets of literals, we have
\[
C\subseteq C^*.
\]
Hence, any interpretation satisfying $C$ also
satisfies $C^*$.

Furthermore, since $C_{i_0}$ is an arbitrary clause
of $S$, its corresponding clause $C_{i_0}^*$ is also
arbitrary in $S^*$. Therefore, the clause set $S^*$
associated with the resulting triangular standard
contradiction is also minimally unsatisfiable.
\end{remark}

\begin{remark}
\label{rem:common-framework}
For each theorem-generation procedure corresponding to
$\Dlt_2$, $\Dlt_3$, $\Dlt_4$, $\Dlt_5$,
$\Dlt_6(i)$, $\Dlt_7(i)$, $\Dlt_8(i)$
($i=n-1,\ldots,2$), and $\Dlt$, given a generating
literal set $\{x_1,\ldots,x_n\}$ whose members have
pairwise distinct predicate symbols and do not involve
equality, the construction begins with a triangular
sub-standard contradiction of a full triangular standard
contradiction. We call this initial structure the
\emph{skeleton}. As established in Sections~6.3--6.10, each such skeleton
corresponds to a minimally unsatisfiable clause set
containing exactly $2n$ literal occurrences.

Excluding the skeleton itself, there are $(n-1)!$
possible positions for placing additional literals
in the triangular interior. According to the
literal-placement principle of Definition~\ref{def:triangular-standard},
an arbitrary nonempty set of literals can then be
placed in the triangular interior (the regions
indicated by the oblique dashed lines in Figs.~\ref{fig:delta2-type-I}-~\ref{fig:general-triangular-mu}).

This construction is the same for
$\Dlt_2$, $\Dlt_3$, $\Dlt_4$, $\Dlt_5$,
$\Dlt_6(i)$, $\Dlt_7(i)$, $\Dlt_8(i)$
($i=n-1,\ldots,2$), and $\Dlt$.
The only difference lies in the construction
of their respective skeletons, namely, the
positions of the literals other than those on
the main boundary.
\end{remark}

\begin{remark}
\label{rem:skeleton-count}
In first-order logic, consider a triangular standard
contradiction generated from the literal set
$\{x_1,\ldots,x_n\}$, whose members have pairwise
distinct predicate symbols and do not involve equality.

Suppose that, whenever a clause $C_i$ contains $x_i$,
the corresponding row contains exactly one occurrence
of $\neg x_i$, for $i=1,\ldots,n$, and that
\[
C_{n+1}=\neg x_n,
\]
as illustrated in Fig.~10.

The associated clause set
\[
S=\{C_{n+1},C_n,\ldots,C_1\}
\]
is minimally unsatisfiable.

For each $x_i$, there are $n-i+1$ possible positions
for $\neg x_i$, where $i=1,\ldots,n$.
Since the choices for different $x_i$ do not
interfere with one another, once the order of the
generating literals $x_1,\ldots,x_n$ on the main
boundary is fixed, the number of such minimally
unsatisfiable clause sets is
\[
\prod_{i=1}^{n}(n-i+1)=n!.
\]

Furthermore, there are $n!$ permutations of the
generating literals on the main boundary.
Consequently, a total of
\[
n!\times n!=(n!)^2
\]
skeletons can be generated. Each skeleton
corresponds to one minimally unsatisfiable
clause set.
\end{remark}

\subsection{Number of Theorems Generated from Triangular
Standard Contradictions}
\label{sec:theorem-count}

In view of Remark~\ref{rem:common-framework},
for convenience, we choose the relatively simple
$\Dlt_2$ skeleton for the following discussion.

For $\Dlt_2$, additional literals can be placed in
the triangular interior of each clause $C_i$,
$i=n+1,\ldots,3$. The number of ways to place at
least one additional literal in $C_i$ is
\[
\sum_{j=1}^{i-2}\binom{i-2}{j}.
\]
Including the case in which no additional literal
is placed, the number of possibilities is
\[
\sum_{j=0}^{i-2}\binom{i-2}{j}.
\]

Since the placement choices for different clauses
are independent, the total number of possible
literal-placement configurations for
$C_{n+1},\ldots,C_3$ is
\[
\prod_{i=3}^{n+1}
\left(
\sum_{j=0}^{i-2}\binom{i-2}{j}
\right).
\]

This includes exactly one configuration in which
no additional literal is placed in any clause.
Since at least one additional literal must be
placed in the triangular interior, the number
of nonempty placement configurations is
\[
\prod_{i=3}^{n+1}
\left(
\sum_{j=0}^{i-2}\binom{i-2}{j}
\right)-1.
\]

For each fixed ordering of the generating literals
$x_1,\ldots,x_n$, these configurations give
different clause sets. As there are $n!$
permutations of the generating literals, the
number of clause sets obtained by adding at
least one interior literal is
\[
n!\left[
\prod_{i=3}^{n+1}
\left(
\sum_{j=0}^{i-2}\binom{i-2}{j}
\right)-1
\right].
\]

Furthermore, each of the $n!$ permutations
determines a $\Dlt_2$ skeleton, and each skeleton
itself corresponds to a minimally unsatisfiable
clause set. Including these skeletons, the
total number of minimally unsatisfiable clause sets is
\[
\begin{aligned}
N_{\mathrm{CS}}
&=
n!\left[
\prod_{i=3}^{n+1}
\left(
\sum_{j=0}^{i-2}\binom{i-2}{j}
\right)-1
\right]+n!\\
&=
n!\prod_{i=3}^{n+1}
\left(
\sum_{j=0}^{i-2}\binom{i-2}{j}
\right).
\end{aligned}
\]

Using the binomial identity, the above counting expression can be simplified using
\[
\sum_{j=0}^{i-2}\binom{i-2}{j}
=2^{i-2},
\]
we obtain
\[
\begin{aligned}
N_{\mathrm{CS}}
&=n!\prod_{i=3}^{n+1}2^{i-2}\\
&=n!\,2^{\sum_{i=3}^{n+1}(i-2)}\\
&=n!\,2^{1+2+\cdots+(n-1)}\\
&=n!\,2^{n(n-1)/2}.
\end{aligned}
\]

Thus, the number of clause sets generated by
the $\Dlt_2$ construction, including the
skeletons and their admissible interior-literal
extensions, is
\[
\boxed{
N_{\mathrm{CS}}=n!\,2^{n(n-1)/2}.
}
\]

Each such clause set is minimally unsatisfiable.
Indeed, the $\Dlt_2$ skeletons are minimally
unsatisfiable, and the admissible placement
of additional interior literals preserves
minimal unsatisfiability, as established by
the triangular standard contradiction
construction and
Theorems~\ref{thm:literal-sat}--\ref{thm:joint-sat}.

Now consider any such minimally unsatisfiable
clause set
\[
S=\{C_{n+1},\ldots,C_1\}.
\]
For every nonempty proper subset
$T\subsetneq S$, the remaining clause set
$S\setminus T$ is satisfiable, and
\[
S\setminus T
\vdash
\neg\bigwedge_{C\in T}C.
\]
By Remark~\ref{rem:multi}, the number of
generated theorems is
\[
N_{\mathrm{T}}
=
\sum_{k=1}^{n}\binom{n+1}{k}.
\]

Applying the binomial theorem again,
\[
\begin{aligned}
N_{\mathrm{T}}
&=\sum_{k=0}^{n+1}\binom{n+1}{k}
  -\binom{n+1}{0}-\binom{n+1}{n+1}\\
&=2^{n+1}-2.
\end{aligned}
\]

Consequently, following the counting
construction of the triangular standard
contradiction methods, the total number
of generated theorems is
\[
\begin{aligned}
N_{\mathrm{total}}
&=
\left[
n!\prod_{i=3}^{n+1}
\left(
\sum_{j=0}^{i-2}\binom{i-2}{j}
\right)
\right]
\left[
\sum_{k=1}^{n}\binom{n+1}{k}
\right]\\
&=
n!\,2^{n(n-1)/2}
\left(2^{n+1}-2\right).
\end{aligned}
\]

Therefore, the original counting expression
has the following equivalent closed form:
\[
\boxed{
N_{\mathrm{total}}
=
n!\,2^{n(n-1)/2}
\left(2^{n+1}-2\right).
}
\]

As described in the original construction,
the same counting framework applies to
$\Dlt_3$, $\Dlt_4$, $\Dlt_5$,
$\Dlt_6(i)$, $\Dlt_7(i)$, $\Dlt_8(i)$
($i=n-1,\ldots,2$), and the general
$\Dlt$ method, with the corresponding
differences lying in their skeleton
constructions.

Every generated entailment follows from
a satisfiable premise set and therefore the principle of
explosion does not occur.

For a given set of generating literals
$\{x_1,\ldots,x_n\}$ of a triangular standard
contradiction in first-order logic, assume that
their predicate symbols are pairwise distinct
and that none of the literals involves the
equality predicate. For each skeleton, the
number of distinct theorems that can be
generated without explosion is
\[
N_{\mathrm{total}}
= n!\,2^{n(n-1)/2}(2^{n+1}-2).
\]

By Remark~6.11, the number of distinct
skeletons that can be generated is
$n!\times n!=(n!)^2$.
Furthermore, a full triangular standard
contradiction generates
$n!(2^{n+1}-2)$ theorems. Consequently, the $\Delta$-class construction
methods can generate a total of
\[
\begin{aligned}
N_{\Delta}
&= (n!)^3\,2^{n(n-1)/2}(2^{n+1}-2)
   + n!(2^{n+1}-2)\\
&= \left[(n!)^3\,2^{n(n-1)/2}+n!\right]
   (2^{n+1}-2)
\end{aligned}
\]
distinct theorems for which no explosion occurs.

\section{Conclusion and Outlook}
\label{sec:conclusion}

\noindent This paper has developed a constructive framework for automated theorem
generation in first-order logic based on triangular standard contradictions.
The central idea is to construct minimally unsatisfiable clause sets with
prescribed triangular structures and to transform these structures into
logical entailments by moving one or more selected clauses to the conclusion
while retaining the remaining clauses as premises. In this way, theorem
generation is grounded automatically in the logical properties of the constructed
clause sets rather than in a separate search for a proof of each generated
candidate.

Building on the previously developed and released $\Delta_1$ automated theorem generator
\cite{xu2025delta1}, we have established a systematic family of
triangular-standard-contradiction constructions, including the specific
$\Delta_1$--$\Delta_8(i)$ families and a general $\Delta$-class construction.
For the corresponding constructions, the required unsatisfiability and
satisfiability properties have been established, thereby guaranteeing the
logical validity of the resulting entailments. In particular, when a clause
is removed from a minimally unsatisfiable clause set, the remaining clause
set is satisfiable. Hence the corresponding entailments does not arise
vacuously from inconsistent premises through the principle of explosion.
For the constructions in which multiple clauses are transferred to the
conclusion, satisfiability of the remaining premise set follows from the
corresponding structural results established in this paper.

The general $\Delta$-class construction further shows that a minimally
unsatisfiable triangular structure can be obtained using $2n$ literals under
the stated conditions. Additional admissible interior literals can be
introduced while preserving minimal unsatisfiability, provided that the
conditions of the construction are maintained. Together with the counting
analysis, these results demonstrate that the proposed framework can
systematically generate large families of distinct, logically valid
theorems from a given generating literal set. The significance of this
result lies not merely in the number of generated formulas, but in the
existence of a unified structural mechanism through which their logical
validity follows by construction.

The proposed approach therefore differs from representative
conjecture-oriented methods, in which candidate statements are generated and
then subjected to proving or checking
\cite{claessen2010,yang2019,sivaraman2022}, and from proof-structure
approaches, in which theorem generation proceeds through the construction
or extraction of proof structures \cite{zhou2024,wernhard2026}. Here, the
primary generative object is a structured minimally unsatisfiable clause
set. Once the required structural conditions have been established, no
separate proof process is required for each generated entailment.
This provides a novel perspective on automated theorem generation
and connects the structural theory of minimal unsatisfiability
\cite{aharoni1986,kullmann2011} with constructive theorem discovery.

This paper introduces a class of methods for generating
theorems in first-order logic. The correctness of the
generated theorems is guaranteed by the underlying
generation mechanisms, requiring no additional
verification, and no explosion occurs. However, how
these theorems, expressed in symbolic form, can be
meaningfully interpreted and effectively applied to
solve practical problems in the natural sciences,
engineering and technology, and social sciences
remains an very important direction for future research. In addition, automated mathematic theory exploration
techniques \cite{mccasland2017}, other theorem-generation methods
\cite{puzis2006,lin2024,wernhard2026}, and learning-based models may form
complementary mechanisms for this purpose.

Another promising
direction is to combine the constructive framework with Neuro-Symbolic and
large-language-model methods \cite{alhessi2026lemmanaid}: symbolic construction can
provide formally guaranteed logical structures, while learned models may
assist in selecting, interpreting, and applying generated theorems in
mathematical or other domain-specific contexts.

Overall, the results establish triangular standard contradictions and
minimal unsatisfiability as a constructive basis for automated theorem
generation. They provide a systematic route from structured contradiction
to valid entailment and lay the theoretical foundation for further work on
the generation, evaluation, and application of formally guaranteed logical
knowledge.

\bibliographystyle{elsarticle-num}
\bibliography{references}

@article{wos1993,
  author  = {Wos, Larry},
  title   = {The problem of automated theorem finding},
  journal = {Journal of Automated Reasoning},
  volume  = {10},
  number  = {1},
  pages   = {137--138},
  year    = {1993}
}

@article{zhang2026,
  author  = {Zhang, J. and Tan, S.},
  title   = {Automated Conjecturing and Theorem Finding: A Survey},
  journal = {Journal of Computer Science and Technology},
  volume    = {46},
  number  = {1},
  pages     = {46--66},
  year    = {2026},
  doi     = {10.1007/s11390-026-6040-0}
}

@inproceedings{yang2019,
  author    = {Yang, W. and Fedyukovich, G. and Gupta, A.},
  title     = {Lemma synthesis for automating induction over algebraic data types},
  booktitle = {Principles and Practice of Constraint Programming – CP 2019, LNCS},
  volume    = {11802},
  pages     = {600--617},
  year      = {2019},
    doi     = {10.1007/978-3-030-30048-7_35}
}

@article{sivaraman2022,
  author  = {Sivaraman, A. and Sanchez-Stern, A. and Chen, B. and Lerner, S. and Millstein, T.},
  title   = {Data-driven lemma synthesis for interactive proofs},
  journal = {Proceedings of the ACM on Programming Languages},
  number  = {143},
  pages   = {505–-531},
  year    = {2022},
  doi     = {10.1145/3563306}
}

@inproceedings{claessen2010,
  author    = {Claessen, K. and Smallbone, N. and Hughes, J.},
  title     = {{QuickSpec}: Guessing formal specifications using testing},
  booktitle = {Tests and Proofs – TAP 2010, LNCS},
  volume    = {6143},
  pages     = {6--21},
  year      = {2010},
    doi     = {10.1007/978-3-642-13977-2_3}
}

@inproceedings{puzis2006,
  author    = {Puzis, Y. and Gao, Y. and Sutcliffe, G.},
  title     = {Automated generation of interesting theorems},
  booktitle = {Proceedings of the 19th International Florida Artificial Intelligence Research Society Conference},
  pages     = {49--54},
  year      = {2006}
}

@inproceedings{lin2024,
  author    = {Lin, X. and Cao, Q. and Huang, Y. and Yang, Z. and Liu, Z. and Li, Z. and Liang, X.},
  title     = {{ATG}: Benchmarking automated theorem generation for generative language models},
  booktitle = {Findings of the Association for Computational Linguistics: NAACL},
  pages     = {4465--4480},
  year      = {2024},
  doi     = {10.18653/v1/2024.findings-naacl.279}
}

@inproceedings{zhou2024,
  author    = {Zhou, J. P. and Wu, Y. and Li, Q. and Grosse, R.},
  title     = {{REFACTOR}: Learning to extract theorems from proofs},
  booktitle = {International Conference on Learning Representations},
  year      = {2024}
}

@misc{alhessi2026lemmanaid,
  title         = {{Lemmanaid}: Neuro-Symbolic Lemma Conjecturing},
  author        = {Yousef Alhessi and S{\'o}lr{\'u}n Halla Einarsd{\'o}ttir
                   and George Granberry and Emily First and Moa Johansson
                   and Sorin Lerner and Nicholas Smallbone},
  year          = {2026},
  eprint        = {2504.04942},
  archivePrefix = {arXiv},
  primaryClass  = {cs.AI}
}

@inproceedings{sheng2025,
  author    = {Sheng, Jiayi and Lyu, Luna and Jin, Jikai and Xia, Tanglin
               and Gu, Alex and Zou, James Y. and Lu, Pan},
  title     = {Solving Inequality Proofs with Large Language Models},
  booktitle = {Advances in Neural Information Processing Systems},
  volume    = {38},
  year      = {2025}
}

@inproceedings{petrov2026brokenmath,
  title     = {{BrokenMath}: A Benchmark for Sycophancy in Theorem Proving with {LLM}s},
  author    = {Petrov, Ivo and Dekoninck, Jasper and Vechev, Martin},
  booktitle = {Proceedings of the 43rd International Conference on Machine Learning},
  series    = {Proceedings of Machine Learning Research},
  volume    = {306},
  pages     = {98430--98458},
  year      = {2026},
  publisher = {PMLR},
  eprint    = {2510.04721},
  archivePrefix = {arXiv},
  primaryClass  = {cs.AI}
}

@article{xu2018,
  author  = {Xu, Y. and Liu, J. and Chen, S. W. and Zhong, X. M. and He, X. X.},
  title   = {Contradiction separation based dynamic multi-clause synergized automated deduction},
  journal = {Information Sciences},
  volume  = {462},
  pages   = {93--113},
  year    = {2018}
}

@article{xu2026,
  author  = {Xu, Y. and Chen, S. W. and Zhong, X. M. and Liu, J. and He, X. X.},
  title   = {Towards multi-clause automated deduction and theorem generation: constructing and applying standard contradictions},
  journal = {Knowledge-Based Systems},
  volume  = {334},
  pages   = {114985},
  year    = {2026}
}

@article{xu2025ext,
  author  = {Xu, Y. and He, X. X. and Chen, S. W. and Liu, J. and Zhong, X. M.},
  title   = {Dynamic deduction by contradiction separation: the standard extension algorithm},
  journal = {arXiv preprint arXiv:2510.08468},
  year    = {2025}
}

@article{xu2025tri,
  author  = {Xu, Y. and Chen, S. W. and Liu, J. and Cao, F. and He, X. X.},
  title   = {Extended triangular method: a generalized algorithm for contradiction separation based automated deduction},
  journal = {arXiv preprint arXiv:2510.10701},
  year    = {2025}
}

@article{xu2025rect,
  author  = {Xu, Y. and Liu, P. Y. and Chen, S. W. and Liu, J.},
  title   = {An automated theorem generator with theoretical foundation based on rectangular standard contradiction},
  journal = {arXiv preprint arXiv:2511.04092},
  year    = {2025}
}

@phdthesis{zeng2025,
  author = {Zeng, G. Y.},
  title  = {Inverse and Parallel Automated Theorem Prover Based on Contradiction Separation},
  school = {Southwest Jiaotong University},
  year   = {2025}
}

@article{zeng2024,
  author  = {Zeng, G. Y. and Chen, S. W. and Liu, J. and Xu, Y. and Liu, P. Y.},
  title   = {A complementary ratio based clause selection method for contradiction separation dynamic deduction},
  journal = {Knowledge-Based Systems},
  volume  = {284},
  pages   = {111238},
  year    = {2024}
}

@phdthesis{liu2024,
  author = {Liu, P. Y.},
  title  = {Study on a First-Order Logic Automated Theorem Prover Based on Contradiction Separation Deduction Rule},
  school = {Southwest Jiaotong University},
  year   = {2024}
}

@article{liu2023a,
  author  = {Liu, P. Y. and Chen, S. W. and Liu, J. and Xu, Y. and Cao, F. and Wu, G. F.},
  title   = {An efficient contradiction separation based automated deduction algorithm for enhancing reasoning capability},
  journal = {Knowledge-Based Systems},
  volume  = {261},
  pages   = {110217},
  year    = {2023}
}

@article{liu2023b,
  author  = {Liu, P. Y. and Xu, Y. and Liu, J. and Chen, S. W. and Cao, F. and Wu, G. F.},
  title   = {Fully reusing clause deduction algorithm based on standard contradiction separation rule},
  journal = {Information Sciences},
  volume  = {622},
  pages   = {337--356},
  year    = {2023}
}

@phdthesis{cao2020,
  author = {Cao, F.},
  title  = {Study on a First-Order Logic Automated Theorem Prover Based on Contradiction Separation Deduction},
  school = {Southwest Jiaotong University},
  year   = {2020}
}

@article{cao2021,
  author  = {Cao, F. and Xu, Y. and Liu, J. and Chen, S. W. and Yi, J.},
  title   = {A multi-clause dynamic deduction algorithm based on standard contradiction separation rule},
  journal = {Information Sciences},
  volume  = {566},
  pages   = {281--299},
  year    = {2021}
}

@article{cao2019,
  author  = {Cao, F. and Xu, Y. and Chen, S. W. and Zhong, J. and Wu, G. F.},
  title   = {A contradiction separation dynamic deduction algorithm based on optimized proof search},
  journal = {International Journal of Computational Intelligence Systems},
  volume  = {12},
  number  = {2},
  pages   = {1245--1254},
  year    = {2019}
}

@book{harrison2009,
  author    = {Harrison, John},
  title     = {Handbook of Practical Logic and Automated Reasoning},
  publisher = {Cambridge University Press},
  address   = {Cambridge},
  year      = {2009}
}

@article{xu2025delta1,
  author  = {Xu, Y. and Guo, H. L. and Chen, S. W. and Liu, J.},
  title   = {{$\Delta_1$}: An automated theorem generator},
  journal = {International Journal of Computational Intelligence Systems},
  volume  = {18},
  pages   = {309},
  year    = {2025}
}

@article{aharoni1986,
  author  = {Aharoni, R. and Linial, N.},
  title   = {Minimal non-two-colorable hypergraphs and minimal unsatisfiable formulas},
  journal = {Journal of Combinatorial Theory, Series A},
  volume  = {43},
  pages   = {196--204},
  year    = {1986}
}

@misc{delta1github,
  author       = {Xu, Y. and Guo, H. L. and Chen, S. W. and Liu, J.},
  title        = {Automated Theorem Generator {$\Delta_1$}},
  year         = {2026},
  month        = jun,
  howpublished = {GitHub release},
  url          = {https://github.com/SWJTU-math/Automated_Theorem_Generator}
}

@misc{delta2github,
  author       = {Xu, Y. and Guo, H. L. and Chen, S. W. and Liu, J.},
  title        = {Automated Theorem Generator {$\Delta_2$}},
  year         = {2026},
  month        = jun,
  howpublished = {GitHub release},
  url          = {https://github.com/SWJTU-math/Automated-Theorem-Generator-2}
}

@misc{delta3github,
  author       = {Xu, Y. and Guo, H. L. and Chen, S. W. and Liu, J.},
  title        = {Automated Theorem Generator {$\Delta_3$}},
  year         = {2026},
  month        = jun,
  howpublished = {GitHub release},
  url          = {https://github.com/SWJTU-math/Automated-Theorem-Generator-3}
}

@misc{delta4github,
  author       = {Xu, Y. and Guo, H. L. and Chen, S. W. and Liu, J.},
  title        = {Automated Theorem Generator {$\Delta_4$}},
  year         = {2026},
  month        = jun,
  howpublished = {GitHub release},
  url          = {https://github.com/SWJTU-math/Automated-Theorem-Generator-4}
}

@inproceedings{wernhard2026,
  author    = {Wernhard, Christoph},
  title     = {Generating Theorems by Generating Proof Structures},
  booktitle = {Automated Reasoning -- IJCAR 2026},
  series    = {Lecture Notes in Computer Science},
  volume    = {16688},
  pages     = {41--61},
  publisher = {Springer},
  year      = {2026}
}

@article{mccasland2017,
  author  = {McCasland, R. L. and Bundy, A. and Smith, P. F.},
  title   = {{MATHsAiD}: Automated mathematical theory exploration},
  journal = {Applied Intelligence},
  volume  = {47},
  pages   = {585--606},
  year    = {2017},
  doi     = {10.1007/s10489-017-0954-8}
}

@article{kullmann2011,
  author  = {Kullmann, Oliver},
  title   = {Constraint Satisfaction Problems in Clausal Form {II}:
             Minimal Unsatisfiability and Conflict Structure},
  journal = {Fundamenta Informaticae},
  volume  = {109},
  number  = {1},
  pages   = {83--119},
  year    = {2011},
  doi     = {10.3233/FI-2011-429}
}

@article{kullmann2016,
  author        = {Oliver Kullmann},
  title         = {Minimal unsatisfiability and deficiency:
                   recent developments},
  journal       = {CoRR},
  volume        = {abs/1610.08575},
  year          = {2016},
  eprint        = {1610.08575},
  archivePrefix = {arXiv},
  primaryClass  = {cs.LO},
  }

\newpage

\appendix

\section{Proof of Theorem~\ref{thm:literal-sat}}\label{app:thm41}
\noindent Let $x=P(t_1,\ldots,t_m)$, where $P$ is an $m$-ary predicate symbol and $t_1,\ldots,t_m$ are terms.

(1) To show that $x$ is satisfiable, take any nonempty domain $D$, for example $D=\{a\}$, and interpret $P$ by $P^D=D^m$. Interpret all function and constant symbols arbitrarily. Thus $P$ is true of every $m$-tuple. Under this interpretation, $P(t_1,\ldots,t_m)$ is true for every variable assignment, so $x$ is satisfiable.

(2) To show that $\neg x$ is satisfiable, take a nonempty domain $D$, again for example $D=\{a\}$, and interpret $P$ by $P^D=\varnothing$. Thus $P$ is false of every $m$-tuple. Since $P$ is not the equality predicate, this interpretation is always permitted. Therefore $P(t_1,\ldots,t_m)$ is false for every assignment and $\neg x$ is true. Hence $\neg x$ is satisfiable. \qed

\section{Proof of Theorem~\ref{thm:joint-sat}}\label{app:thm42}
(1) Construct a model $M$ in which all $x_i$ are simultaneously true. Since the literals have pairwise distinct predicate symbols, the interpretation of each predicate can be defined independently. Take any nonempty domain, for example the singleton $\{a\}$. Interpret every individual constant as $a$ and every function symbol as the constant function with value $a$. For each predicate symbol $P$ occurring in a literal $x_i$, define $P^M$ as follows. If $x_i$ is the positive literal $P(t_1,\ldots,t_m)$, include the tuple $(t_1^M,\ldots,t_m^M)$ in $P^M$; other tuples may be assigned arbitrarily. If $x_i$ is the negative literal $\neg P(t_1,\ldots,t_m)$, exclude $(t_1^M,\ldots,t_m^M)$ from $P^M$; other tuples may again be assigned arbitrarily. Because different $x_i$ use different predicate symbols, these requirements do not conflict. Hence every $x_i$ is true in $M$, so $\{x_1,\ldots,x_n\}$ is satisfiable.

(2) For each $k$, write $G_k=x_k\vee H_k$, where $H_k$ is an arbitrary disjunction. In the model $M$ constructed in part (1), $x_k$ is true, so $G_k$ is true. Hence $\{G_1,\ldots,G_n\}$ is satisfiable. \qed

\section{Proof of Theorem~\ref{thm:mu-theorem-generation}}\label{app:thm43}
Since $T\neq\varnothing$, choose any clause $C\in T$. Because $S$ is
minimally unsatisfiable,
\[
S\setminus\{C\}
\]
is satisfiable. Moreover,
\[
S\setminus T
\subseteq
S\setminus\{C\}.
\]
Any subset of a satisfiable clause set is satisfiable. Hence
$S\setminus T$ is satisfiable.

It remains to establish the entailment. Suppose, for contradiction, that
there exists a model $M$ such that
\[
M\models S\setminus T
\]
and
\[
M\not\models
\neg\bigwedge_{C\in T}C.
\]
Then
\[
M\models
\bigwedge_{C\in T}C.
\]
Thus $M$ satisfies every clause in $T$ as well as every clause
in $S\setminus T$, and therefore
\[
M\models S,
\]
contradicting the unsatisfiability of $S$. Hence
\[
S\setminus T
\models
\neg\bigwedge_{C\in T}C.
\]
The corresponding syntactic consequence follows from the completeness
theorem for first-order logic.
\qed

\section{Proof of Lemma~\ref{lem:clause-deletion}}\label{app:lem51}

\medskip
\noindent We consider two cases:
$y=P(\mathbf{t})$ and
$y=\neg P(\mathbf{t})$.

\medskip
\noindent\textbf{Case 1:}
$y=P(\mathbf{t})$, where $P$ is a predicate
symbol and $\mathbf{t}$ is a tuple of terms
(a $k$-tuple of terms if $P$ is a $k$-ary
predicate).

By assumption, the predicate symbol $P$
does not occur in any clause of $S'$,
either positively as $P(\ldots)$ or
negatively as $\neg P(\ldots)$.

\medskip
\noindent\textbf{(I) Suppose that $S'$ is
satisfiable. We prove that $S$ is satisfiable.}

Let $\mathcal{M}=(D,J)$ be a model of $S'$,
where $D$ is the domain and $J$ is an
interpretation. Thus, $\mathcal{M}$ satisfies
the universal closure of every clause in $S'$.

Construct a new model
$\mathcal{M}'=(D',J')$ as follows:
\begin{enumerate}
    \item The domain remains unchanged:
    \[
        D'=D.
    \]

    \item The interpretations of all symbols
    other than $P$, including function
    symbols, constants, and other predicate
    symbols, are identical to those in
    $\mathcal{M}$.

    \item The predicate $P$ is interpreted
    as universally true in $\mathcal{M}'$.
    For a unary predicate, $P$ is true
    for every individual in $D$. More
    generally, for a $k$-ary predicate,
    its interpretation is defined as
    \[
        P^{\mathcal{M}'}=D^k,
    \]
    so that $P$ is true for every
    $k$-tuple of elements of $D$.
\end{enumerate}

We now prove that $\mathcal{M}'$ satisfies $S$.

Consider any clause $E\in S'$.
Since no literal in $E$ contains the predicate
symbol $P$, and $\mathcal{M}'$ agrees with
$\mathcal{M}$ on the interpretations of all
symbols occurring in $E$, the truth value
of $E$ under $\mathcal{M}'$ is identical
to that under $\mathcal{M}$ for every
variable assignment. Since
$\mathcal{M}\models S'$, it follows that
$\mathcal{M}'\models E$. Therefore,
\[
    \mathcal{M}'\models S'.
\]

Next, consider the clause $C$.
The tuple $\mathbf{t}$ consists of terms
formed from constants, function symbols,
and variables, and contains no predicate
symbols. Consequently, changing the
interpretation of $P$ does not affect
the interpretation of $\mathbf{t}$.
Thus, $\mathbf{t}$ has exactly the same
value in $\mathcal{M}'$ as in $\mathcal{M}$,
regardless of the variable assignment.

Since $P$ is universally true in
$\mathcal{M}'$, the atomic formula
$P(\mathbf{t})$ is true under every
variable assignment. The clause $C$
is a disjunction of the form
\[
    C=P(\mathbf{t})\lor C_0,
\]
where $C_0$ denotes the disjunction
of the remaining literals, if any.
Since one of its disjuncts,
$P(\mathbf{t})$, is true, the entire
clause $C$ is true under every
variable assignment. Hence,
\[
    \mathcal{M}'\models C.
\]

Therefore,
\[
    \mathcal{M}'\models S,
\]
and consequently $S$ is satisfiable.

\medskip
\noindent\textbf{(II) Suppose that $S$ is
satisfiable. We prove that $S'$ is satisfiable.}

Since $S'\subseteq S$, every model
satisfying $S$ necessarily satisfies $S'$.
Therefore, $S'$ is satisfiable.

Combining (I) and (II), we obtain
\[
    S\text{ is satisfiable}
    \iff
    S'\text{ is satisfiable}.
\]
Thus, the clause $C$ can be removed
without changing satisfiability.

\medskip
\noindent\textbf{Case 2:}
$y=\neg P(\mathbf{t})$.

In the construction used in Case 1,
replace the interpretation of $P$ as
universally true with an interpretation
under which $P$ is universally false.
Specifically, define
\[
    P^{\mathcal{M}'}=\varnothing.
\]
For a unary predicate, this means that
$P$ is false for every individual in $D$;
for a $k$-ary predicate, it means that
$P$ is false for every $k$-tuple
of elements of $D$.

All other interpretations remain unchanged.
The remainder of the argument is identical
to that in Case 1. In particular, every
clause in $S'$ remains satisfied, while
the literal $\neg P(\mathbf{t})$ is true
under every variable assignment, ensuring
that $C$ is satisfied.

Consequently,
\[
    S\text{ is satisfiable}
    \iff
    S'\text{ is satisfiable}.
\]
Hence, removing the clause $C$ does not
change satisfiability.

The lemma therefore holds in both cases.
\qed

\section{Proof of Lemma~\ref{lem:add}}\label{app:lem52}

\medskip
\noindent Let $\mathcal{M}=(D,I)$ be a model of $S$.
Denote the predicate symbol of $y$ by $P$,
and suppose that $P$ has arity $n$. Thus,
\[
    y=P(t_1,\ldots,t_n)
    \quad\text{or}\quad
    y=\neg P(t_1,\ldots,t_n),
\]
where each $t_i$ is a term.

Construct a new model
$\mathcal{M}'=(D,I')$ over the same domain
$D$, where $I'$ agrees with $I$ on all
function symbols, constants, and predicate
symbols other than $P$. The interpretation
of $P$ is defined as follows:
\begin{itemize}
    \item If $y=P(t_1,\ldots,t_n)$, let
    \[
        P^{\mathcal{M}'}=D^n,
    \]
    i.e., $P$ is interpreted as the
    universal relation on $D$.

    \item If $y=\neg P(t_1,\ldots,t_n)$, let
    \[
        P^{\mathcal{M}'}=\varnothing,
    \]
    i.e., $P$ is interpreted as the
    empty relation on $D$.
\end{itemize}

We now show that
\[
    \mathcal{M}'\models S\cup\{y\}.
\]

\begin{enumerate}
    \item For any clause $C\in S$,
    the predicate symbol $P$ does not
    occur in $C$ by assumption.
    Therefore, all symbols occurring
    in $C$ have identical interpretations
    under $I'$ and $I$. Consequently,
    $C$ has the same truth value in
    $\mathcal{M}'$ as in $\mathcal{M}$
    under every variable assignment.
    Since $\mathcal{M}\models S$,
    it follows that
    \[
        \mathcal{M}'\models S.
    \]

    \item For the unit clause $y$,
    consider the following two cases.

    If $y=P(t_1,\ldots,t_n)$, then
    \[
        (t_1,\ldots,t_n)^{\mathcal{M}',s}
        \in D^n=P^{\mathcal{M}'}
    \]
    for every variable assignment $s$.
    Hence,
    \[
        \mathcal{M}'\models
        P(t_1,\ldots,t_n).
    \]

    If $y=\neg P(t_1,\ldots,t_n)$,
    then $P^{\mathcal{M}'}=\varnothing$,
    so no $n$-tuple belongs to
    $P^{\mathcal{M}'}$. Therefore,
    for every variable assignment $s$,
    \[
        \mathcal{M}'\models
        \neg P(t_1,\ldots,t_n).
    \]

    Thus, in either case,
    \[
        \mathcal{M}'\models y.
    \]
\end{enumerate}

Combining the above results, we obtain
\[
    \mathcal{M}'\models S\cup\{y\}.
\]
Therefore, $S\cup\{y\}$ is satisfiable.
 \qed

\section{Proof of Theorem~\ref{thm:general}}\label{app:thm51}

\textbf{Part I.} We proceed by mathematical induction on the number
of generating literals (corresponding to the number
of rows in Fig.~\ref{fig:general-triangular-mu}, from $1$ to $n$).
Consider a triangular standard contradiction
(as shown in Fig.~\ref{fig:general-triangular-mu}), whose rows, from top
to bottom, correspond to $k=1,\ldots,n$.

\medskip
\noindent\textbf{1) Base case: $k=1$.}

When $k=1$, let
\[
S=\{C_{n+1},C_n\},
\qquad
C_{n+1}=\neg x_n,\quad C_n=x_n.
\]
Clearly, $S$ is a minimally unsatisfiable
clause set.

\medskip
\noindent\textbf{2) Inductive hypothesis.}

Suppose that, for $1<k\leq n$,
\[
S=\{C_{n+1},C_n,\ldots,C_{i_0+1},
C_{i_0},C_{i_0-1},\ldots,C_1\}
\]
is a minimally unsatisfiable clause set.

\medskip
\noindent\textbf{3) Inductive step: $k=n+1$.}

We prove that
\[
S^+=\{C_{n+1},C_n,\ldots,C_{i_0+1},
C_{i_0}^+,C_{i_0-1},\ldots,C_1,C\},
\]
where
\[
C=y,\qquad
C_{i_0}^+=C_{i_0}\lor\neg y,
\]
is a minimally unsatisfiable clause set.

First, observe that, according to the
structure of a triangular standard
contradiction, the generating literal
set at this stage is
\[
\{y,x_1,\ldots,x_n\},
\]
where the predicate symbols are pairwise
distinct and none is the equality predicate.

Denote the resulting triangular standard
contradiction by $\Pi(S^+)$. Here, the names
of the sets of literals corresponding to
the individual clauses remain the same
as the clause names.

Then
\[
\prod_{i=1}^{n+1} C_i \times C
\]
is a subsumption-free clause set in the
full triangular standard contradiction
constructed from the generating literal
set $\{y,x_1,\ldots,x_n\}$.

Therefore, $S^+$ is a triangular standard
contradiction contained in this full
triangular standard contradiction.
Consequently, $S^+$ is an unsatisfiable
clause set.

We now prove that $S^+$ is minimally
unsatisfiable.

\medskip
\noindent\textbf{(i)}
Consider removing $C$ from $S^+$.

By Lemma~\ref{lem:clause-deletion}, the clause $C_{i_0}^+$
in $S^+$ is a redundant clause after
$C$ has been removed. Thus,
$C_{i_0}^+$ can also be removed without
changing satisfiability.

The remaining clause set is
\[
S^+\setminus\{C_{i_0}^+, C\}
=
\{C_{n+1},C_n,\ldots,C_{i_0+1},
C_{i_0-1},\ldots,C_1\}.
\]

At this point, the remaining clause set
contains $n$ rows. By the inductive
hypothesis, $S^+\setminus\{C\}$ is
satisfiable.

\medskip
\noindent\textbf{(ii)}
Consider removing $C_{i_0}^+$ from $S^+$.

By Lemma~\ref{lem:clause-deletion}, the clause $C$ becomes
redundant in $S^+$ after $C_{i_0}^+$
has been removed. Hence, $C$ can also
be removed without changing satisfiability.

The remaining clause set is
\[
S^+\setminus\{C,C_{i_0}^+\}
=
\{C_{n+1},C_n,\ldots,C_{i_0+1},
C_{i_0-1},\ldots,C_1\}.
\]

The remaining clause set contains $n$
rows. By the inductive hypothesis,
$S^+\setminus\{C_{i_0}^+\}$ is
satisfiable.

\medskip
\noindent\textbf{(iii)}
Consider removing any other clause
$C_i$ from $S^+$, where
\[
C_i\neq C_{i_0}^+,\qquad C_i\neq C,
\]
and
\[
i=n+1,\ldots,i_0+1,i_0-1,\ldots,1.
\]

Assign $y$ the truth value true.
Then $C_{i_0}^+$ has the same truth
value as $C_{i_0}$, since
\[
C_{i_0}^+=C_{i_0}\lor\neg y
\]
and $\neg y$ is false under this
assignment.

By the inductive hypothesis, removing
$C_i$ from the clause set corresponding
to the $k=n$ rows yields a satisfiable
clause set. That is,
\[
\{C_{n+1},\ldots,C_{i+1},
C_{i-1},\ldots,C_1\}
\]
is satisfiable.

Furthermore, by Lemma~\ref{lem:add},
\[
\{C_{n+1},\ldots,C_{i+1},
C_{i-1},\ldots,C_1,C\}
\]
is also satisfiable.

Consequently,
$S^+\setminus\{C_i\}$ is satisfiable.

\medskip
Combining (i), (ii), and (iii), we
conclude that $S^+$ is a minimally
unsatisfiable clause set.

\medskip
\noindent\textbf{Part II.}

For the generating literal set
$\{x_1,\ldots,x_n\}$ of a triangular
standard contradiction $S$, there are
$n$ rows.

For each $x_i$ ($i=1,\ldots,n$),
among the literals occurring in the
same row as $x_i$, apart from $x_i$
itself, there is only $\neg x_i$.
Thus, each row contains exactly
two literals, $x_i$ and $\neg x_i$
($i=1,\ldots,n$), as shown in
Fig.~\ref{fig:general-triangular-mu}.

Therefore, a triangular standard
contradiction constructed in this way
contains a total of $2n$ literals
and $n+1$ clauses
$\{C_{n+1},\ldots,C_1\}$.

Since $S$ is a minimally unsatisfiable
clause set, suppose that any occurrence
of a literal $x_{i_0}$ is removed.
By Lemma~\ref{lem:clause-deletion}, the clause containing
$\neg x_{i_0}$ can then be removed
without changing satisfiability.
This contradicts the minimal
unsatisfiability of $S$.

Similarly, suppose that any occurrence
of a literal $\neg x_{i_0}$ is removed.
By Lemma~\ref{lem:clause-deletion}, the clause containing
$x_{i_0}$ can then be removed without
changing satisfiability. This again
contradicts the minimal unsatisfiability
of $S$.

Therefore, the number of literal
occurrences in $S$ cannot be less
than $2n$.

Hence, the minimum number of literal
occurrences in $S$ is $2n$.
\qed

\end{document}